\documentclass{article}

\usepackage[preprint]{neurips_2025}

\usepackage[utf8]{inputenc} 
\usepackage[T1]{fontenc}    
\usepackage{hyperref}       
\usepackage{url}            
\usepackage{booktabs}       
\usepackage{amsfonts}       
\usepackage{nicefrac}       
\usepackage{microtype}      
\usepackage{xcolor}         
\usepackage{amsmath}
\usepackage{amssymb}
\usepackage{mathtools}
\usepackage{amsthm}
\usepackage{subcaption}
\usepackage{multirow}
\usepackage{graphicx}
\usepackage{pifont}
\usepackage{algorithm}
\usepackage{algpseudocode}
\usepackage{enumitem}
\usepackage{wrapfig}

\title{Wavelet Mixture of Experts for Time Series Forecasting}

\author{%
  Zheng Zhou \\
  Shanghai University of Engineering Science\\
  Shanghai 201620, China\\
  \texttt{M320123332@sues.edu.cn} \\
  \And
  Yu-Jie Xiong \\
  Shanghai University of Engineering Science\\
  Shanghai 201620, China\\
  \texttt{xiong@sues.edu.cn} \\
  \And
  Jia-Chen Zhang \\
  Shanghai University of Engineering Science\\
  Shanghai 201620, China\\
  \texttt{m325123603@sues.edu.cn} \\
  \And
  Chun-Ming Xia \\
  Shanghai University of Engineering Science\\
  Shanghai 201620, China\\
  \texttt{cmxia@sues.edu.cn} \\
  \And
  Xi-Jiong Xie \\
  Ningbo University \\
  Ningbo 315211, China \\
  \texttt{xiexijiong@nbu.edu.cn} \\
}

\begin{document}
\maketitle

\begin{abstract}
The field of time series forecasting is rapidly advancing, with recent large-scale Transformers and lightweight Multilayer Perceptron (MLP) models showing strong predictive performance. However, conventional Transformer models are often hindered by their large number of parameters and their limited ability to capture non-stationary features in data through smoothing. Similarly, MLP models struggle to manage multi-channel dependencies effectively. To address these limitations, we propose a novel, lightweight time series prediction model, WaveTS-B. This model combines wavelet transforms with MLP to capture both periodic and non-stationary characteristics of data in the wavelet domain. Building on this foundation, we propose a channel clustering strategy that incorporates a Mixture of Experts (MoE) framework, utilizing a gating mechanism and expert network to handle multi-channel dependencies efficiently. We propose WaveTS-M, an advanced model tailored for multi-channel time series prediction. Empirical evaluation across eight real-world time series datasets demonstrates that our WaveTS series models achieve state-of-the-art (SOTA) performance with significantly fewer parameters. Notably, WaveTS-M shows substantial improvements on multi-channel datasets, highlighting its effectiveness.
\end{abstract}

\section{Introduction}
Time series forecasting is vital across various domains, as precise predictions enable more detailed planning. With the evolution of deep learning technologies~\cite{lecun2015deep}, a wide array of tools has become available for time series analysis, including Recurrent Neural Networks (RNNs)~\cite{zhang2023robust}, Graph Neural Networks (GNNs)~\cite{huang2023crossgnn}, and Transformers~\cite{liu2024itransformer}. Many time series exhibit inherent periodicity (single or multiple cycles), such as the 24-hour cycle observed in electrical energy data, which persists over a long period of time. In addition, the non-stationarity and channel correlation of data, such as weather patterns that may change over time, and the entanglement and interference of multiple influencing factors by random factors, pose challenges to long-term forecasting~\cite{10454253,10531218,10414287,10709894,9684678,10674771}. 
Capturing these long-term dependencies often requires extensive historical data, increasing the model's complexity and parameter count, thereby lengthening both training and inference times. This issue is pronounced in popular Transformer-based models, which can contain millions of parameters and become significantly less efficient as input lengths grow, limiting their practical application~\cite{wu2021autoformer,zhou2022fedformer}.

\begin{figure}[ht]
	\centering
	\begin{subfigure}{0.5\textwidth}
		\includegraphics[width=\linewidth]{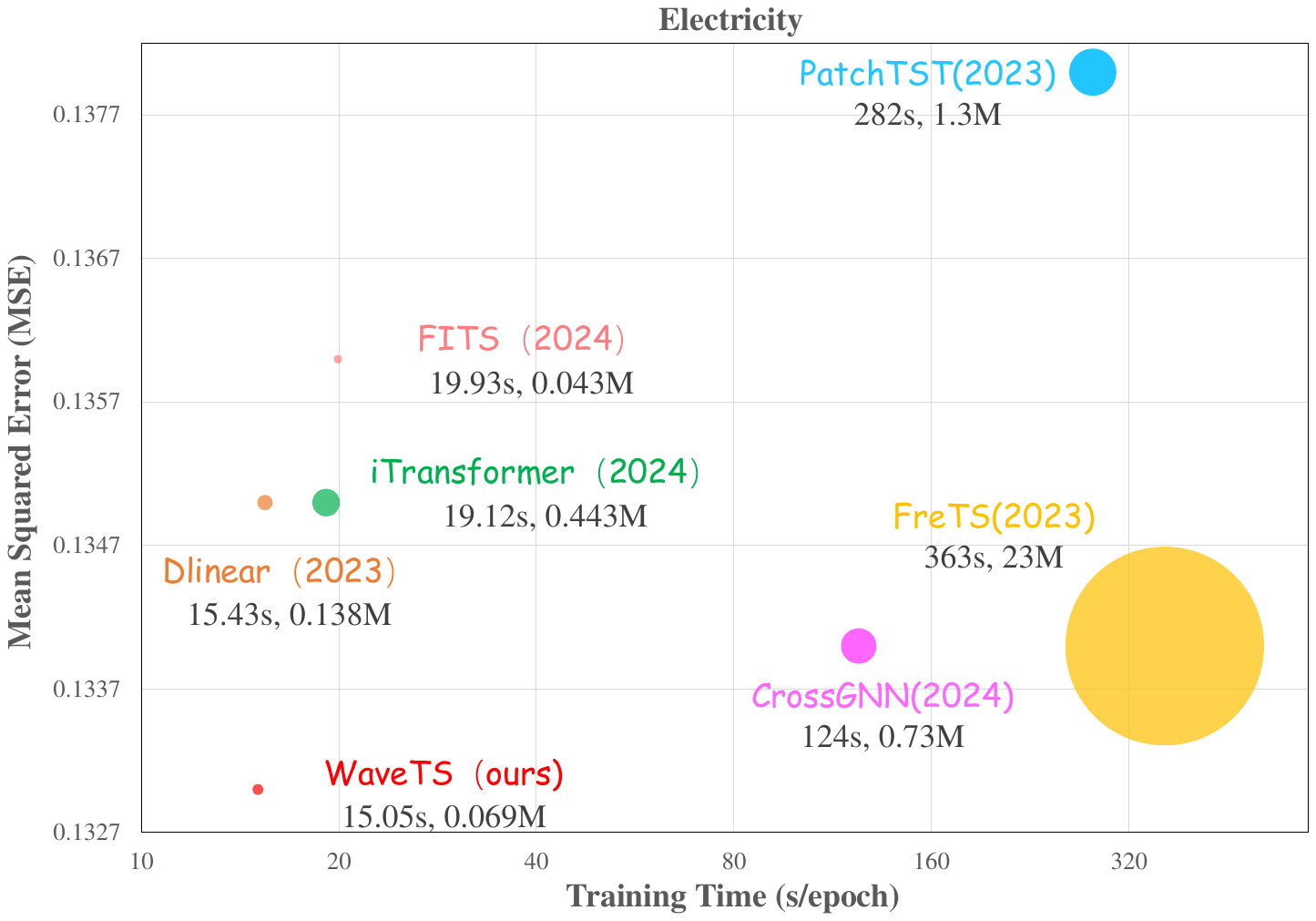}
	\end{subfigure}\hfill
	\begin{subfigure}{0.5\textwidth}
		\includegraphics[width=\linewidth]{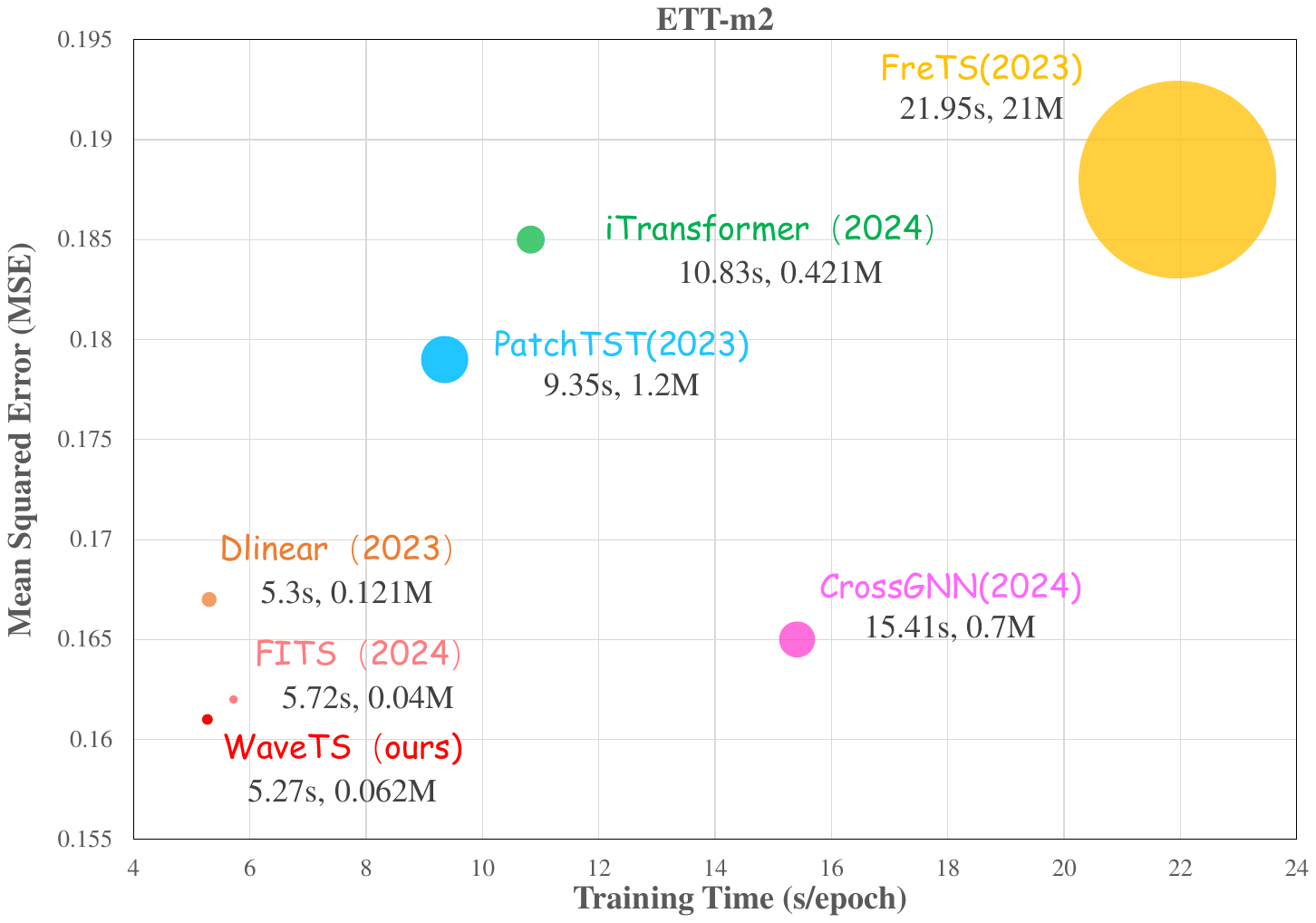}
	\end{subfigure}
	\caption{Comparison of performance, efficiency, and parameter quantity between our model and other mainstream models on Electricity and ETT-h2 datasets. WaveTS-B demonstrate excellent performance and exhibits significant efficiency advantages compared to larger models, achieving state-of-the-art (SOTA) performance with a lightweight structure. Input length is $L=720$ and prediction length is $S=96$.}
\label{Bubble Diagram}
\end{figure}

In this work, we propose the WaveTS series models, WaveTS-B and WaveTS-M, innovative and efficient models for time series forecasting, utilizing time-frequency analysis in the wavelet domain.WaveTS-M's architecture unfolds in three stages: the initial transformation of the sequence through orthogonal learnable high-pass and low-pass filters, and the downsampling of the resultant high and low-frequency components to shorten the input sequence. Subsequently, the predicted results are weighted and combined using gate-controlled networks and expert networks~\cite{jacobs1991adaptive,Shazeer2017OutrageouslyLN,xue2022go}. The final prediction is obtained by combining these time-domain signals. Our approach offers three primary benefits: (i) The wavelet transform splits the time series into approximate and detailed representations, achieving a reversible, lossless downsampling that retains most of the critical information, such as periodicity, in the low-frequency components, while the high-frequency parts capture disruptions and noise. This efficiency in data utilization makes wavelet transforms highly effective. (ii) The use of matrix multiplication in the model is streamlined to mere additions and subtractions of sequences, significantly reducing the time needed to process time series. This streamlined approach lays the groundwork for developing lightweight models that do not sacrifice analytical depth for efficiency. (iii) Using a channel clustering strategy to address the challenge of multi-channel data correlations not only prevents a significant increase in model complexity but also enhances the expressive capacity of the model.

Despite being a relatively straightforward model, WaveTS series models consistently attain SOTA performance on publicly available real-world datasets spanning multiple domains. As shown in Figure \ref{Bubble Diagram}, WaveTS-B demonstrates advantages in predictive performance and efficiency. Overall, our contributions are summarized as follows:

\begin{itemize}
	\item We propose WaveTS-B, a fundamental yet robust model for time series prediction that employs specialized filters to execute wavelet transforms. This transformation decomposes the time series into high-frequency and low-frequency components, allowing the model to concentrate on the aspects most significant for prediction. Subsequently, the model leverages the nonlinear transformation capabilities of an MLP to process these components. This dual approach enables effective extraction and utilization of both periodic and non-stationary features of the data, enhancing the model’s predictive performance.
	
	\item We propose a channel clustering strategy that integrates a MoE framework to enhance the WaveTS-B model, resulting in the advanced multi-channel time series prediction model, WaveTS-M. This strategy assigns weights to each channel's significance across different experts through a gating network, where the allocation of each channel's weight is not fixed but probabilistic. Such flexibility ensures that information from each channel can be utilized by multiple experts, allowing for dynamic channel combinations by different experts based on their relative contributions in varying contexts. This adaptive approach significantly improves the model's versatility and efficiency in managing diverse and complex data scenarios.

    \item Our proposed WaveTS series methods achieve SOTA predictive performance on real-world datasets and have parameter advantages.
\end{itemize}

\section{Related work}
Numerous deep learning approaches have been developed for time series forecasting, each with its own strengths and challenges. RNN-based models, such as LSTNet~\cite{lai2018modeling}, struggle with increasing computational costs and error accumulation as the prediction horizon grows. CNN-based methods, including SCINet~\cite{liu2022scinet} and TCN~\cite{lea2017temporal}, face difficulties in capturing long-range global dependencies. Recent innovations have partially addressed these challenges by integrating causal convolution with large kernels, improving the models' ability to capture broader temporal patterns.
Transformers, including Informer~\cite{zhou2021informer} and Reformer~\cite{kitaev2020reformer}, initially demonstrated promising results but were later outperformed by simpler models in certain contexts. More recent work, such as PatchTST~\cite{nie2023a}, has successfully applied concepts from Vision Transformers to time series forecasting, achieving significant improvements. Additionally, iTransformer~\cite{liu2024itransformer} introduced novel methods for modeling data-token relationships, leading to impressive performance in time series prediction tasks.
Decomposing data is a key step in deep learning, especially with the growing focus on seasonal decomposition in time series analysis. RobustSTL~\cite{wen2019robuststl} offers a novel time series decomposition algorithm that robustly extracts trends by solving regression issues with sparse regularization and minimum absolute deviation loss. Autoformer~\cite{wu2021autoformer} innovates by replacing the Transformer's self-attention with Fast Fourier Transform (FFT) for more efficient sequence-level connections and delayed aggregation.
Recently, there has been a shift towards the frequency domain in research, as scientists convert complex time-domain data into more clear frequency-domain data, facilitating easier learning of time series characteristics. FiLM~\cite{zhou2022film} employs the Fourier transform to reduce noise. FreTS~\cite{yi2024frequency} introduces a framework that learns channel and time dependencies in the frequency domain. FITS~\cite{xu2024fits} offers a streamlined linear model that transforms time series forecasting into interpolation training in the complex frequency domain, using linear layers designed to enable amplitude scaling and phase shifting.
Predicting time series in the wavelet domain is gaining interest as a novel approach. FEDformer~\cite{zhou2022fedformer} merges the Discrete Fourier Transform (DFT) with Transformer techniques to process features in the frequency domain. CoST~\cite{woo2022cost} utilizes DFT's intermediate layer for frequency transformation. WFTNet~\cite{liu2023wftnet} introduces a hybrid model that employs both Fourier and wavelet transforms to capture global and local patterns effectively, achieving notable performance.

\section{Method}
\label{Method}
In this section, we offer a comprehensive and detailed exposition of the proposed WaveTS series models. These models involve a pivotal transformation from the time domain to the wavelet domain, which facilitates in-depth analysis of time series data from both time and frequency perspectives. This transformation allows for the effective dissection of the data's intrinsic structures and patterns that may not be apparent in the time domain alone. Additionally, the models employ a channel clustering strategy to adeptly manage the correlations among multiple channels in time series data. This strategy ensures that our models can accurately capture and utilize the complex interdependencies that exist within multi-dimensional data sets, enhancing predictive accuracy and model robustness.

\noindent\textbf{Research Problem.} Multivariate time series data commonly comprise several interrelated variables, necessitating advancements in forecasting methodologies. We consider the dataset $X = {x_1, \dots, x_L} \in \mathbb{R}^{L \times N}$, where $x_l$ captures the multivariate observations at the $l$-th time step. The goal is to forecast future values for $S$ subsequent time steps, expressed as $Y = {y_1, \dots, y_S} \in \mathbb{R}^{S \times N}$. Here, $X_{l,:}$ indicates the data observed at time $l$, and $X_{:,n}$ represents the complete time series for the $n$-th variable.

\subsection{Preliminary: wavelet transform}
The wavelet transform plays a crucial role in decomposing time series data into approximate and detailed components, facilitating the simultaneous capture of temporal and frequency characteristics. This property renders the wavelet transform particularly effective for examining the distributional properties of time series data, proving immensely beneficial in diverse analytical scenarios. Specifically, the application of the Discrete Wavelet Transform (DWT) \cite{heil1989} to a time series of length $L$ typically yields two sequences of features, each extending to $L/2 + C$, where $C$ denotes additional coefficients contingent on the selected wavelet basis \cite{daubechies1988orthonormal}.

The mathematical formulation of this decomposition is given by:
\begin{equation} S(t) = A_{j_0}(t) + \sum_{j > j_0} D_j(t), \end{equation}
where $S(t)$ represents the time series. $A_{j_0}(t)$ and $D_j(t)$, denoting the approximation and detail components at scale $j_0$ and higher scales $j$, respectively, are computed through:
\begin{equation} A_{j_0}(t) = \sum_{k} A_{j_0,k} \varphi_{j_0,k}(t), D_j(t) = \sum_{k} D_{j,k} \psi_{j,k}(t), \end{equation}
where coefficients $A_{j_0,k}$ and $D_{j,k}$ are obtained by:
\begin{equation} A_{j_0,k} = \langle S, \varphi_{j_0,k} \rangle, D_{j,k} = \langle S, \psi_{j,k} \rangle.\end{equation}
The scaling and wavelet functions are defined as follows:
\begin{equation} \varphi_{j,k}(t) = 2^{\frac{j}{2}} \varphi(2^j t - k), \psi_{j,k}(t) = 2^{\frac{j}{2}} \psi(2^j t - k).\end{equation}
These equations illustrate the operations of scaling and translation applied to the signal. Following Fourier theory, temporal compression equates to spectral expansion and upward frequency shift, described by:
\begin{equation} \mathcal{F}(S(at)) = \frac{1}{|a|} \mathcal{F}\left(\frac{\omega}{a}\right), \end{equation}
where $\mathcal{F}$ represents the Fourier transform, and $a$ is the scaling factor. This relationship implies that compressing the time sequence by factor $a$ not only diminishes its amplitude by $\frac{1}{a}$ but also scales the frequency from $\omega$ to $\frac{\omega}{a}$.

\subsection{WaveTS}

\begin{figure*}[th]
	\centering
	\includegraphics[width=\textwidth]{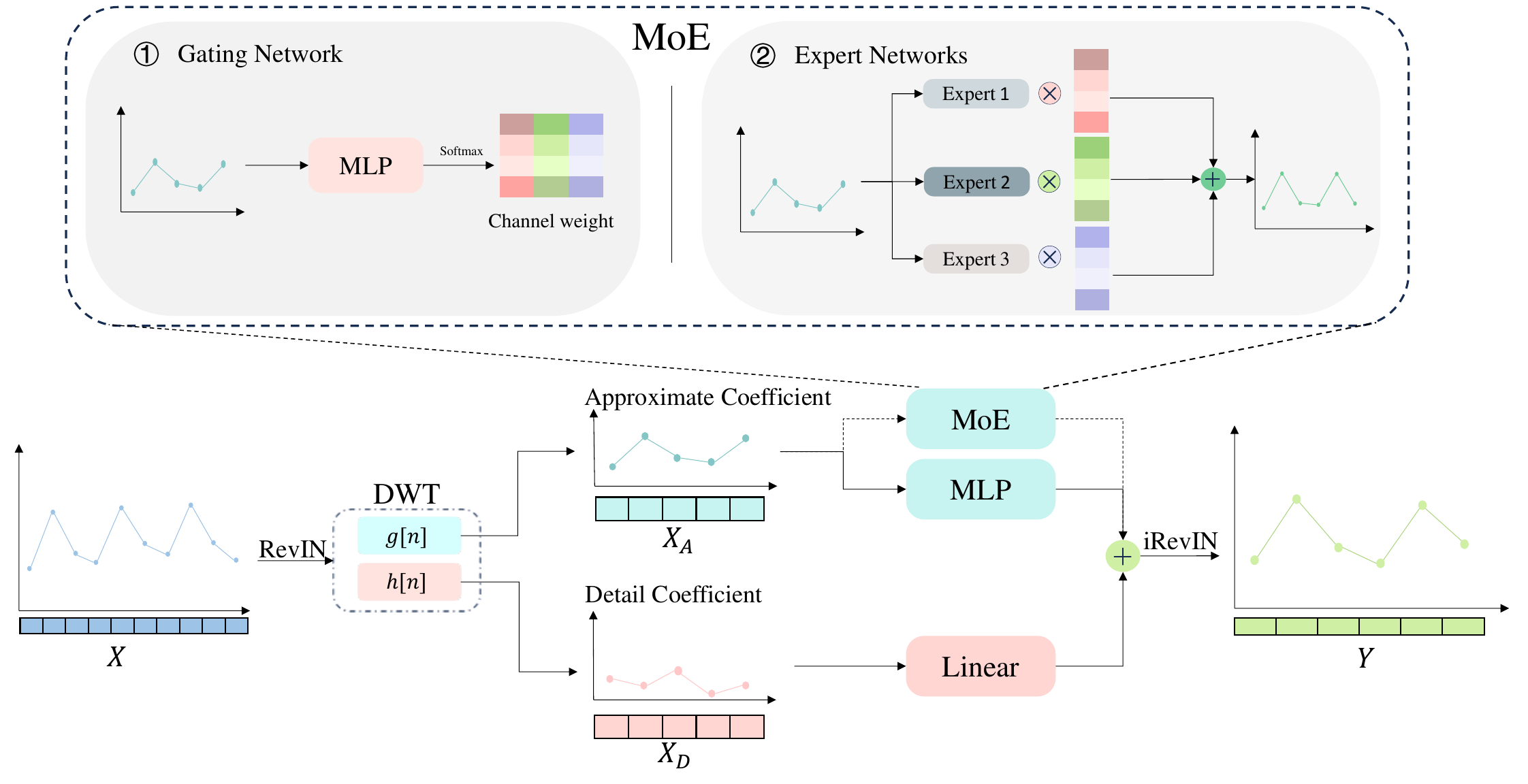}
	\caption{The pipeline of WaveTS series models. We commence with the application of RevIN for reversible instance normalization on time-series data. This is followed by a discrete wavelet transform, employing high-pass and low-pass filters, to decompose the data into approximate coefficients (representing low-frequency components) and detail coefficients (representing high-frequency components). For the WaveTS-B model, these low-frequency components are processed and predicted using an MLP. On the other hand, the WaveTS-M model takes a more nuanced approach by utilizing a MoE module. Within this module, gating networks assign channel weights to the low-frequency components, facilitating the generation of independent predictions from each expert. These predictions are then aggregated, taking into account the weights assigned to each expert's output, to derive the composite prediction for the low-frequency elements. The final forecast is produced by merging the predictions for both the low-frequency and high-frequency components. This consolidated result is then subjected to inverse instance normalization using iRevIN, restoring the data to its original scale and distribution. This comprehensive methodology ensures a robust prediction framework capable of handling intricate data dynamics in time-series analysis.}
	\label{pipline}
\end{figure*}

The wavelet transform is applied iteratively to decompose the time series into $j+1$ feature sequences via $j$ levels of decomposition. The first sequence represents the approximation component, capturing the low-frequency aspects of the signal, while the subsequent $j$ sequences detail the signal's characteristics across various high-frequency bands.
For the wavelet transformation in WaveTS, we employ a set of simple orthogonal filter coefficients, encapsulated within a transformation matrix, as expressed below:

\begin{equation}
	Matrix=\begin{bmatrix}
		\alpha&\alpha \\
		\alpha&-\alpha
	\end{bmatrix}
	=\begin{bmatrix}
		g[k] &
		h[k]
	\end{bmatrix},
\end{equation}
where $k$ denotes the index variable traversing the filter coefficients, with $g[k]$ representing low-pass, and $h[k]$ high-pass filter coefficients. The Discrete Wavelet Transform (DWT) in WaveTS is computed through matrix convolution: 
\begin{equation} 
	DWT(X) = X \ast Matrix = \sum_{k=0}^{K-1} [X_A, X_D].
\end{equation} 
The outputs of the DWT, denoted as $X_A$ and $X_D$, are defined as follows: 
\begin{equation} 
	X_A = \sum_{k=0}^{K-1} X[2n-k] \cdot g[k], X_D = \sum_{k=0}^{K-1} X[2n-k] \cdot h[k],
\end{equation} 
where $X_A$ captures the low-frequency content typically associated with higher energy, and $X_D$ captures the high-frequency content, generally considered to contain noise and minimal relevant information. The output of Expert networks (wavelet MLPs) $E(X)$ is represented by the following formula:
\begin{equation}
	E(X) = \mathbf{W}_e \text{ReLU}(\mathbf{W}_h \mathbf{X} + \mathbf{b}_h) + \mathbf{b}_e,
\end{equation}
where $\mathbf{W}_e$, $\mathbf{W}_h$, $\mathbf{b}_h$ and $\mathbf{b}_e$ are the weights and biases of the expert network.

Typically, we deploy the WaveTS-B model for long-term prediction tasks, employing channel weight-sharing techniques to treat each channel equally. This approach achieves a balance between performance and computational efficiency. However, when handling multi-channel data, such as traffic flow data comprising 862 channels, using a weight-sharing approach may fail to highlight key channels, thus not fully leveraging important data characteristics. If a channel independence approach were adopted, where each channel is assigned its own network, it would substantially increase both the complexity and computational demands of the model. To address these issues, we employ the WaveTS-M model, which utilizes channel clustering to make independent predictions for each category while sharing weights within categories. This method allows the model to more effectively capture and utilize the diversity of input data, thereby enhancing performance in various scenarios. The pipeline of our model is shown in Figure \ref{pipline}. The WaveTS processing pipeline begins with the normalization of the time series data using Reversible Instance Normalization (RevIN) \cite{kim2021reversible}. This step ensures that the data maintains a consistent scale and distribution throughout the analysis. Following normalization, the time series undergoes wavelet transform to separate it into low-frequency (approximation coefficients) and high-frequency (detail coefficients) components. These components represent different aspects of the underlying signal dynamics. The high-frequency part undergoes a linear transformation to prepare it for further processing. Depending on the specific model variant used—WaveTS-B for a more straightforward approach or WaveTS-M for handling more complex data scenarios—the transformed low-frequency data is then processed either through an MLP or a MoE module. Each method is designed to effectively capture and leverage the intrinsic patterns within the low-frequency data. Following the processing of the high-frequency components, the outputs from either MLP or MoE are fused with the high-frequency data. This fusion integrates the distinct characteristics of both frequency ranges, ensuring a comprehensive representation of the time series. The process concludes with a final normalization step, where the combined prediction results are normalized using inverse RevIN. This step is crucial for returning the data to its original form, ensuring the predictions are accurately aligned with the original data's scale and distribution.

MoE consists of a learnable gate network $G(X)$ and expert $E(X)$. The weight obtained by each channel through the gating network determines its importance among different experts:
\begin{equation}
	G(X) = \text{Softmax}(\mathbf{W}_g \mathbf{X} + \mathbf{b}_g),
\end{equation}
where $\mathbf{W}_g$ and $\mathbf{b}_g$ are the weights and biases of the gating network. 
The result of MoE module is obtained by weighting the outputs of all experts with the weights of the gate network outputs:
\begin{equation}
MoE(X) = \sum G(X)E(X).
\end{equation}
The final prediction result is composed of the MoE module and the high-frequency prediction results combined:
\begin{equation}
Y = MoE(X_A) + \delta Linear(X_D).
\end{equation}
In order to provide a clearer explanation of the principle of WaveTS and improve the readability and reproducibility of the model, we present the complete algorithm in Appendix~\ref{algorithm}.

\section{Experiments}
We undertake comprehensive experimental analyses using eight real-world time series benchmarks to assess the performance and computational efficiency of the WaveTS series models. These experiments are designed to benchmark WaveTS against SOTA methods in terms of forecasting accuracy and resource utilization.

\noindent\textbf{Datasets.} All datasets are publicly available and widely used real-world datasets from different fields, mainly including Electricity, Traffic, Weather, Exchange rates, and ETT used by Autoformer~\cite{wu2021autoformer}. The details are shown in Appendix~\ref{database}.

\noindent\textbf{Baselines.} To thoroughly evaluate the performance and efficiency of WaveTS series models, we compare them against a range of SOTA time series forecasting models. The selected baselines encompass a diverse set of architectures, including both time-domain and frequency-domain models, as well as traditional machine learning and deep learning approaches. Specifically, the models compared are PatchTST~\cite{nie2023a}, DLinear~\cite{zeng2023transformers}, iTransformer~\cite{liu2024itransformer}, FITS~\cite{xu2024fits}, FreTS~\cite{yi2024frequency}, Reformer~\cite{kitaev2020reformer}, Informer~\cite{zhou2021informer}, and CrossGNN~\cite{huang2023crossgnn}. 

\noindent\textbf{Implementation details.} Following the experimental setup of FITS~\cite{xu2024fits}, we configure the length of the input sequence to be $T=720$, while the prediction sequence length is set to $S \in \left\{ 96, 192, 336, 720 \right\}$. To mitigate the risk of information leakage, normalization is applied independently to each time slot rather than the entire dataset, ensuring that future values do not influence past data points.
The evaluation metrics used to assess model performance are Mean Squared Error (MSE) \cite{allen1971mean} and Mean Absolute Error (MAE) \cite{willmott2005advantages}, providing a robust measure of both error magnitude and consistency. All experiments are conducted on a single NVIDIA RTX 3090 GPU 24GB and implemented using PyTorch \cite{paszke2019pytorch}.
\label{Implementation details}

\subsection{Model comparison}
In this section, we compare the performance of WaveTS series models and other SOTA models in long-term and short-term time series prediction tasks. The efficiency comparison includes parameter quantity and training time.

\subsubsection{Long-term forecasting results}
\label{subsub1}
\begin{table*}[th]	
	\caption{Long-term forecasting comparison. The best results are in \textbf{bold} and the second best results are in \underline{underlined}. Use the mean to balance the differences in different prediction lengths and count the best and second results.}
	\scriptsize
	\centering
	\resizebox{\textwidth}{!}{\setlength{\tabcolsep}{1mm}{
			\begin{tabular}{c c c c|c c|c c|c c|c c|c c|c c|c c|c c}
				\hline
				\multicolumn{2}{c}{Models}&\multicolumn{2}{|c|}{\textbf{WaveTS-B(ours)}}&\multicolumn{2}{c|}{DLinear}&\multicolumn{2}{c|}{FITS}&\multicolumn{2}{c|}{FreTS}&\multicolumn{2}{c|}{iTransformer}&\multicolumn{2}{c|}{Reformer}&\multicolumn{2}{c|}{Informer}&\multicolumn{2}{c|}{PatchTST}&\multicolumn{2}{c}{CrossGNN}\\
				\cline{1-20}
				\multicolumn{2}{c|}{Metric}&MSE&MAE&MSE&MAE&MSE&MAE&MSE&MAE&MSE&MAE&MSE&MAE&MSE&MAE&MSE&MAE&MSE&MAE\\
				\hline
				\multicolumn{1}{c|}{\multirow{5}{*}{\rotatebox{90}{Exchange}}}&\multicolumn{1}{c|}{96}&\textbf{0.083}&\underline{0.203}&\underline{0.087}&0.213&0.088&0.208&0.525&0.531&0.118&0.253&1.117&0.902&1.104&0.866&0.124&0.251&\textbf{0.083}&\textbf{0.198}\\
				\multicolumn{1}{c|}{}&\multicolumn{1}{c|}{192}&\textbf{0.174}&\textbf{0.297}&0.196&0.337&0.181&0.302&0.958&0.731&0.232&0.355&1.158&0.913&1.175&0.875&0.287&0.390&\underline{0.183}&\underline{0.30}\\
				\multicolumn{1}{c|}{}&\multicolumn{1}{c|}{336}&\underline{0.338}&0.424&\textbf{0.269}&\textbf{0.387}&\underline{0.338}&\underline{0.418}&1.100&0.765&0.434&0.489&1.291&0.959&1.297&0.960&0.702&0.586&0.348&0.425\\
				\multicolumn{1}{c|}{}&\multicolumn{1}{c|}{720}&1.025&0.762&\textbf{0.946}&\underline{0.738}&\underline{0.964}&\textbf{0.733}&2.518&1.188&1.050&0.750&1.530&1.051&1.165&0.908&1.371&0.867&1.221&0.828\\
				\cline{2-20}
				\multicolumn{1}{c|}{}&\multicolumn{1}{c|}{Avg}&0.405&0.420&\textbf{0.374}&\underline{0.418}&\underline{0.392}&\textbf{0.415}&1.275&0.803&0.459&0.463&1.274&0.956&1.185&0.902&0.621&0.523&0.458&0.437\\
				\hline
				\multicolumn{1}{c|}{\multirow{5}{*}{\rotatebox{90}{weather}}}&\multicolumn{1}{c|}{96}&0.167&0.220&0.170&0.230&0.168&0.223&\underline{0.154}&\underline{0.214}&0.169&0.222&0.398&0.421&0.217&0.294&\textbf{0.149}&\textbf{0.205}&0.162&0.218\\
				\multicolumn{1}{c|}{}&\multicolumn{1}{c|}{192}&0.210&\underline{0.257}&0.220&0.280&0.211&0.258&\underline{0.199}&0.261&0.211&0.258&0.651&0.562&0.303&0.353&\underline{0.199}&\textbf{0.248}&\textbf{0.197}&0.250\\
				\multicolumn{1}{c|}{}&\multicolumn{1}{c|}{336}&0.256&\underline{0.293}&0.258&0.310&0.257&0.294&\underline{0.253}&0.309&0.273&0.302&0.641&0.554&0.485&0.477&\textbf{0.248}&\textbf{0.291}&0.249&0.294\\
				\multicolumn{1}{c|}{}&\multicolumn{1}{c|}{720}&0.319&\underline{0.338}&0.321&0.364&0.322&0.341&0.329&0.366&0.325&0.343&0.713&0.606&0.732&0.614&\underline{0.317}&\textbf{0.335}&\textbf{0.313}&0.337\\
				\cline{2-20}
				\multicolumn{1}{c|}{}&\multicolumn{1}{c|}{Avg}&0.238&\underline{0.277}&0.242&0.296&0.239&\underline{0.279}&\underline{0.233}&0.287&0.244&0.281&0.600&0.535&0.434&0.434&\textbf{0.228}&\textbf{0.269}&0.230&0.274\\
				\hline
				\multicolumn{1}{c|}{\multirow{5}{*}{\rotatebox{90}{Electricity}}}&\multicolumn{1}{c|}{96}&\underline{0.133}&\textbf{0.228}&0.135&0.234&0.136&0.235&0.134&0.234&0.135&\underline{0.232}&0.294&0.377&0.529&0.531&\textbf{0.130}&\textbf{0.228}&0.134&0.231\\
				\multicolumn{1}{c|}{}&\multicolumn{1}{c|}{192}&\textbf{0.148}&\textbf{0.242}&0.150&0.249&0.151&0.248&0.151&0.249&0.153&0.251&0.298&0.377&0.567&0.541&\underline{0.149}&\underline{0.245}&\underline{0.149}&\underline{0.245}\\
				\multicolumn{1}{c|}{}&\multicolumn{1}{c|}{336}&\textbf{0.164}&\textbf{0.258}&\textbf{0.164}&\underline{0.262}&0.167&0.264&0.422&0.292&0.167&0.266&0.348&0.418&0.542&0.554&\underline{0.166}&\textbf{0.258}&0.165&0.261\\
				\multicolumn{1}{c|}{}&\multicolumn{1}{c|}{720}&0.203&\textbf{0.291}&\underline{0.199}&0.297&0.205&0.296&0.455&0.312&\textbf{0.196}&\underline{0.292}&0.334&0.406&0.547&0.556&0.210&0.298&0.204&0.294\\
				\cline{2-20}
				\multicolumn{1}{c|}{}&\multicolumn{1}{c|}{Avg}&\textbf{0.162}&\textbf{0.254}&\textbf{0.162}&0.260&0.164&0.260&0.290&0.271&\underline{0.163}&\underline{0.260}&0.318&0.394&0.451&0.485&0.163&0.257&0.163&0.257\\
				\hline
				\multicolumn{1}{c|}{\multirow{5}{*}{\rotatebox{90}{Traffic}}}&\multicolumn{1}{c|}{96}&\underline{0.377}&\textbf{0.265}&0.387&\underline{0.274}&0.390&\underline{0.274}&0.388&0.390&\textbf{0.371}&0.275&0.663&0.362&0.845&0.487&0.382&0.278&0.394&0.283\\
				\multicolumn{1}{c|}{}&\multicolumn{1}{c|}{192}&\underline{0.390}&\textbf{0.272}&0.400&0.281&0.400&\underline{0.277}&0.412&0.292&\textbf{0.382}&0.278&0.696&0.378&0.897&0.500&\underline{0.396}&0.282&0.407&0.288\\
				\multicolumn{1}{c|}{}&\multicolumn{1}{c|}{336}&\underline{0.403}&\textbf{0.275}&0.412&0.287&0.412&\underline{0.281}&0.422&0.299&\textbf{0.398}&0.285&0.698&0.376&1.463&0.819&\underline{0.404}&0.288&0.441&0.295\\
				\multicolumn{1}{c|}{}&\multicolumn{1}{c|}{720}&\underline{0.442}&\underline{0.294}&0.452&0.295&0.450&0.301&0.455&0.312&\textbf{0.428}&\textbf{0.286}&0.722&0.98&0.618&0.934&0.445&0.296&0.474&0.311\\
				\cline{2-20}
				\multicolumn{1}{c|}{}&\multicolumn{1}{c|}{Avg}&\underline{0.403}&\textbf{0.276}&0.412&0.284&0.413&0.283&0.419&0.323&\textbf{0.394}&\underline{0.281}&0.694&0.378&1.22&0.685&0.406&0.286&0.429&0.294\\
				\hline
				\multicolumn{1}{c|}{\multirow{5}{*}{\rotatebox{90}{ETTh1}}}&\multicolumn{1}{c|}{96}&\textbf{0.377}&\textbf{0.400}&0.385&0.410&\textbf{0.377}&\textbf{0.400}&0.480&0.480&0.404&0.432&1.006&0.773&1.248&0.874&0.379&0.410&\underline{0.38}&\underline{0.409}\\
				\multicolumn{1}{c|}{}&\multicolumn{1}{c|}{192}&0.421&\underline{0.427}&0.427&0.437&\underline{0.412}&\textbf{0.421}&0.552&0.530&0.465&0.473&1.025&0.779&1.246&0.870&0.415&0.432&\textbf{0.408}&\underline{0.427}\\
				\multicolumn{1}{c|}{}&\multicolumn{1}{c|}{336}&0.452&0.446&0.479&0.478&\underline{0.430}&\textbf{0.436}&0.586&0.549&0.501&0.499&1.181&0.821&1.355&0.864&\textbf{0.424}&0.440&0.434&\underline{0.442}\\
				\multicolumn{1}{c|}{}&\multicolumn{1}{c|}{720}&0.470&0.480&0.527&0.533&\textbf{0.428}&\textbf{0.452}&0.645&0.577&0.601&0.563&1.126&0.818&1.297&0.885&\underline{0.447}&\underline{0.469}&0.454&0.471\\
				\cline{2-20}
				\multicolumn{1}{c|}{}&\multicolumn{1}{c|}{Avg}&0.430&0.441&0.454&0.464&\textbf{0.411}&\textbf{0.427}&0.565&0.534&0.492&0.491&1.084&0.797&1.286&0.873&\underline{0.416}&\underline{0.437}&0.419&\underline{0.437}\\
				\hline
				\multicolumn{1}{c|}{\multirow{5}{*}{\rotatebox{90}{ETTh2}}}&\multicolumn{1}{c|}{96}&\textbf{0.270}&\textbf{0.335}&0.296&0.362&\underline{0.271}&\underline{0.336}&0.428&0.454&0.368&0.393&2.774&1.306&3.023&1.44&0.274&0.337&0.280&0.352\\
				\multicolumn{1}{c|}{}&\multicolumn{1}{c|}{192}&\underline{0.336}&\underline{0.378}&0.345&0.394&\textbf{0.331}&\textbf{0.374}&0.603&0.540&0.437&0.438&4.740&1.683&4.546&1.654&0.339&\underline{0.378}&0.342&0.389\\
				\multicolumn{1}{c|}{}&\multicolumn{1}{c|}{336}&\underline{0.358}&0.399&0.455&0.460&\textbf{0.354}&\underline{0.395}&0.532&0.508&0.453&0.455&4.275&1.656&3.780&1.536&0.367&\textbf{0.384}&0.364&0.407\\
				\multicolumn{1}{c|}{}&\multicolumn{1}{c|}{720}&0.385&0.430&0.782&0.621&\textbf{0.378}&\textbf{0.422}&1.050&0.733&0.440&0.466&3.335&1.437&4.495&1.825&\underline{0.384}&\underline{0.427}&0.406&0.444\\
				\cline{2-20}
				\multicolumn{1}{c|}{}&\multicolumn{1}{c|}{Avg}&\underline{0.337}&\underline{0.385}&0.469&0.459&\textbf{0.333}&\textbf{0.381}&0.654&0.558&0.424&0.438&3.781&1.520&3.961&1.613&0.341&0.381&0.348&0.398\\
				\hline
				\multicolumn{1}{c|}{\multirow{5}{*}{\rotatebox{90}{ETTm1}}}&\multicolumn{1}{c|}{96}&\underline{0.300}&\textbf{0.347}&0.318&0.366&0.312&0.354&0.334&0.382&0.332&0.382&0.652&0.593&0.836&0.678&\textbf{0.294}&\underline{0.348}&\underline{0.306}&\underline{0.349}\\
				\multicolumn{1}{c|}{}&\multicolumn{1}{c|}{192}&\underline{0.337}&\textbf{0.368}&0.350&0.383&0.338&\underline{0.369}&0.365&0.397&0.364&0.400&0.780&0.644&0.833&0.670&\textbf{0.334}&0.373&0.342&0.370\\
				\multicolumn{1}{c|}{}&\multicolumn{1}{c|}{336}&0.371&\underline{0.388}&0.375&0.396&\underline{0.367}&\textbf{0.385}&0.415&0.431&0.394&0.415&0.950&0.723&1.067&0.801&\textbf{0.360}&0.392&0.373&0.390\\
				\multicolumn{1}{c|}{}&\multicolumn{1}{c|}{720}&\underline{0.417}&\underline{0.416}&0.427&0.427&\textbf{0.415}&\textbf{0.412}&0.497&0.488&0.447&0.446&1.114&0.803&1.413&0.964&\textbf{0.415}&0.418&0.419&0.417\\
				\cline{2-20}
				\multicolumn{1}{c|}{}&\multicolumn{1}{c|}{Avg}&\underline{0.356}&\textbf{0.379}&0.367&0.393&0.358&\underline{0.380}&0.402&0.424&0.384&0.410&0.874&0.690&1.037&0.778&\textbf{0.350}&0.382&0.360&0.381\\
				\hline
				\multicolumn{1}{c|}{\multirow{5}{*}{\rotatebox{90}{ETTm2}}}&\multicolumn{1}{c|}{96}&\textbf{0.161}&\textbf{0.251}&0.167&0.259&\underline{0.162}&\underline{0.253}&0.188&0.278&0.185&0.268&0.753&0.678&0.512&0.547&0.164&0.254&0.165&0.257\\
				\multicolumn{1}{c|}{}&\multicolumn{1}{c|}{192}&\textbf{0.216}&\textbf{0.290}&0.237&0.316&\underline{0.217}&\underline{0.292}&0.250&0.316&0.269&0.328&1.114&0.826&1.535&0.956&0.221&\underline{0.292}&0.221&0.296\\
				\multicolumn{1}{c|}{}&\multicolumn{1}{c|}{336}&\underline{0.270}&\underline{0.327}&0.282&0.342&\textbf{0.268}&\textbf{0.325}&0.312&0.353&0.334&0.362&2.218&1.158&2.171&1.117&0.277&0.328&0.272&0.334\\
				\multicolumn{1}{c|}{}&\multicolumn{1}{c|}{720}&\textbf{0.350}&\textbf{0.378}&0.389&0.416&\underline{0.360}&\textbf{0.378}&0.391&0.415&0.413&0.415&2.766&1.254&6.218&1.946&0.367&\underline{0.379}&0.358&0.396\\
				\cline{2-20}
				\multicolumn{1}{c|}{}&\multicolumn{1}{c|}{Avg}&\textbf{0.249}&\textbf{0.311}&0.268&0.333&\underline{0.251}&\underline{0.312}&0.285&0.340&0.300&0.343&1.712&0.979&2.609&1.141&0.257&0.313&0.254&0.320\\
				\hline
				\multicolumn{2}{c|}{$1^{st}$Count}&\multicolumn{2}{c|}{\textbf{30}}&\multicolumn{2}{c|}{6}&\multicolumn{2}{c|}{\underline{23}}&\multicolumn{2}{c|}{0}&\multicolumn{2}{c|}{7}&\multicolumn{2}{c|}{0}&\multicolumn{2}{c|}{0}&\multicolumn{2}{c|}{18}&\multicolumn{2}{c}{5}\\
				\hline
				\multicolumn{2}{c|}{$Avg$~$1^{st}$Count}&\multicolumn{2}{c|}{\textbf{6}}&\multicolumn{2}{c|}{2}&\multicolumn{2}{c|}{\underline{5}}&\multicolumn{2}{c|}{0}& \multicolumn{2}{c|}{1}&\multicolumn{2}{c|}{2}&\multicolumn{2}{c|}{0}&\multicolumn{2}{c|}{3}&\multicolumn{2}{c}{0}\\
				\hline
	\end{tabular}}}
	\label{main result}
\end{table*}

The results of long-term time series prediction are presented in Table \ref{main result}, with a fixed input length of $T=720$ and prediction length of $S\in\left \{ 96,192,336,720\right \} $. It is observable that our model has achieved commendable results across all datasets, leading with a first-place count of $30$, surpassing other models, and securing the top spot in average predictive performance on eight datasets. These accomplishments underscore our model’s status at the SOTA level for predictive performance. This success powerfully demonstrates the superiority and potential of using wavelet transform to decompose time series. 

\begin{wraptable}{r}{0.55\textwidth}
	\scriptsize
	\centering
    \caption{Optimization results of WaveTS-M on multi-channel datasets. (Input length $L=720$ and prediction length $S\in {96,192,336,720}$)}
		\begin{tabular}{c c c c|c c}
				\hline
				\multicolumn{2}{c}{Models}&\multicolumn{2}{|c|}{WaveTS-M}&\multicolumn{2}{c}{WaveTS-B}\\
                \hline
				\multicolumn{2}{c|}{Metric}&MSE&MAE&MSE&MAE\\
				\hline
				\multicolumn{1}{c|}{\multirow{4}{*}{\rotatebox{90}{Electricity}}}&\multicolumn{1}{c|}{96}&\textbf{0.130}&\textbf{0.224}&0.133&0.228\\
				\multicolumn{1}{c|}{}&\multicolumn{1}{c|}{192}&\textbf{0.148}&\textbf{0.241}&\textbf{0.148}&0.242\\
				\multicolumn{1}{c|}{}&\multicolumn{1}{c|}{336}&\textbf{0.164}&\textbf{0.258}&\textbf{0.164}&\textbf{0.258}\\
				\multicolumn{1}{c|}{}&\multicolumn{1}{c|}{720}&\textbf{0.201}&\textbf{0.290}&0.203&0.291\\
				\hline
                \multicolumn{1}{c|}{\multirow{4}{*}{\rotatebox{90}{Traffic}}}&\multicolumn{1}{c|}{96}&\textbf{0.362}&\textbf{0.256}&0.377&0.265\\
				\multicolumn{1}{c|}{}&\multicolumn{1}{c|}{192}&\textbf{0.374}&\textbf{0.261}&0.390&0.272\\
				\multicolumn{1}{c|}{}&\multicolumn{1}{c|}{336}&\textbf{0.388}&\textbf{0.268}&0.403&0.275\\
				\multicolumn{1}{c|}{}&\multicolumn{1}{c|}{720}&\textbf{0.430}&\textbf{0.291}&0.442&0.294\\
				\hline
	  \end{tabular}
	\label{Ablation moe}
\end{wraptable}

\subsubsection{Optimization results of WaveTS-M on multi-channel datasets}
The use of a MoE-based channel clustering strategy enhances the model's ability to manage multi-channel dependencies in time series data. As shown in Table \ref{Ablation moe}, this approach significantly improves the model’s predictive performance on multi-channel datasets. The channel clustering employs a gating mechanism where the weight of each channel is not fixed, but probabilistically assigned, allowing information from a single channel to be utilized by multiple experts. This flexibility enables the model to dynamically adjust resources, prioritizing more important channels for specific tasks. By adapting the channel weights according to their contributions in different contexts, the model can effectively capture and leverage the diversity of input data, improving overall performance.

\subsubsection{Short-term forecasting results}
\label{subsub2}
\begin{table*}[th]
    \caption{Short-term forecasting comparison. The best results are in \textbf{bold} and the second best are \underline{underlined}. (Using $S = 96$ as prediction length and input length $T\in\left\{96, 192, 336, 720\right\}$)}
    \scriptsize
    \renewcommand{\arraystretch}{1.2}
    \centering
    \resizebox{\textwidth}{!}{
        \begin{tabular}{c c| c c |c c |c c |c c |c c |c c |c c |c c|c}
            \hline
            \multicolumn{2}{c}{Models}&\multicolumn{2}{|c}{ETTh1}&\multicolumn{2}{|c}{ETTh2}&\multicolumn{2}{|c}{ETTm1}&\multicolumn{2}{|c}{ETTm2}&\multicolumn{2}{|c}{Exchange}&\multicolumn{2}{|c}{Weather}&\multicolumn{2}{|c}{Electricity}&\multicolumn{2}{|c}{Traffic}&\multicolumn{1}{|c}{$1^{st}$Count}\\
            \cline{1-19}
            \multicolumn{2}{c|}{Metric}&MSE&MAE&MSE&MAE&MSE&MAE&MSE&MAE&MSE&MAE&MSE&MAE&MSE&MAE&MSE&MAE&\\
            \cline{1-18}
            \multirow{4}{*}{\rotatebox{90}{\textbf{WaveTS}}}&
            \multicolumn{1}{|c|}{96}&\textbf{0.385}&\underline{0.394}&\textbf{0.289}&\textbf{0.338}&0.351&\textbf{0.372}&\textbf{0.182}&\textbf{0.265}&\underline{0.082}&\underline{0.200}&\underline{0.193}&\underline{0.233}&0.197&\underline{0.273}&\underline{0.645}&\underline{0.383}&\\
            &\multicolumn{1}{|c|}{192}&\textbf{0.379}&\textbf{0.394}&\textbf{0.283}&\textbf{0.338}&\textbf{0.309}&\underline{0.346}&\textbf{0.171}&\textbf{0.255}&\underline{0.085}&\underline{0.203}&\underline{0.185}&\underline{0.229}&0.153&\underline{0.245}&\underline{0.450}&\underline{0.298}&\textbf{28}\\
            &\multicolumn{1}{|c|}{336}&0.379&0.400&\textbf{0.275}&\textbf{0.337}&0.305&\underline{0.346}&\textbf{0.164}&\textbf{0.253}&\underline{0.089}&\textbf{0.208}&0.172&\underline{0.225}&\underline{0.140}&\underline{0.236}&\underline{0.411}&\underline{0.282}&\\
            &\multicolumn{1}{|c|}{720}&\textbf{0.380}&\underline{0.403}&\textbf{0.270}&\textbf{0.335}&\textbf{0.309}&\underline{0.352}&\textbf{0.162}&\textbf{0.253}&\underline{0.090}&\underline{0.211}&\textbf{0.168}&\textbf{0.221}&\textbf{0.133}&\textbf{0.230}&\underline{0.386}&\underline{0.270}&\\
            \hline
            \multirow{4}{*}{\rotatebox{90}{DLinear}}&
            \multicolumn{1}{|c|}{96}&0.386&0.400&0.326&0.381&\underline{0.345}&0.374&0.187&0.281&\textbf{0.077}&\textbf{0.199}&0.197&0.257&\underline{0.194}&0.276&0.648&0.396&\\
            &\multicolumn{1}{|c|}{192}&0.384&0.400&0.289&0.348&\underline{0.310}&0.349&0.173&0.266&\textbf{0.078}&\textbf{0.200}&0.186&0.248&\underline{0.152}&0.247&0.451&0.303&5\\
            &\multicolumn{1}{|c|}{336}&\textbf{0.373}&\underline{0.369}&\underline{0.280}&\underline{0.343}&\underline{0.303}&\underline{0.346}&\underline{0.318}&0.366&0.089&0.215&0.179&0.248&\underline{0.140}&0.238&0.412&0.287&\\
            &\multicolumn{1}{|c|}{720}&0.385&0.410&0.296&0.362&0.167&0.260&0.167&0.259&0.179&0.230&0.170&0.230&\underline{0.135}&0.234&0.387&0.274&\\
            \hline
            \multirow{4}{*}{\rotatebox{90}{FITS}}&
            \multicolumn{1}{|c|}{96}&\textbf{0.385}&\textbf{0.392}&\underline{0.290}&\textbf{0.338}&0.353&0.374&\textbf{0.182}&\textbf{0.265}&0.083&0.201&0.195&0.235&0.199&0.278&0.650&0.388&\\
            &\multicolumn{1}{|c|}{192}&\underline{0.380}&\underline{0.395}&\underline{0.285}&\underline{0.339}&\textbf{0.309}&\textbf{0.345}&\underline{0.172}&\underline{0.257}&0.09&0.211&0.186&0.230&0.157&0.254&0.453&0.301&20\\
            &\multicolumn{1}{|c|}{336}&\textbf{0.374}&\textbf{0.395}&\textbf{0.275}&\textbf{0.337}&\textbf{0.301}&\textbf{0.344}&\underline{0.166}&\underline{0.255}&0.087&\textbf{0.208}&\underline{0.174}&\underline{0.225}&0.146&0.246&0.414&0.286&\\
            &\multicolumn{1}{|c|}{720}&\textbf{0.380}&\textbf{0.402}&\underline{0.271}&\underline{0.336}&\textbf{0.309}&\textbf{0.351}&\underline{0.163}&\underline{0.254}&\textbf{0.088}&\textbf{0.209}&\underline{0.169}&0.223&0.141&0.243&0.394&0.280&\\
            \hline
            \multirow{4}{*}{\rotatebox{90}{iTrans}}&
            \multicolumn{1}{|c|}{96}&0.392&0.412&0.299&0.348&\textbf{0.335}&\textbf{0.372}&0.185&0.268&0.086&0.206&\textbf{0.178}&\textbf{0.216}&\textbf{0.154}&\textbf{0.247}&\textbf{0.421}&\textbf{0.290}&\\
            &\multicolumn{1}{|c|}{192}&0.397&0.417&0.307&0.358&0.323&0.368&0.189&0.273&0.089&0.212&\textbf{0.170}&\textbf{0.214}&\textbf{0.137}&\textbf{0.233}&\textbf{0.378}&\textbf{0.271}&\underline{22}\\
            &\multicolumn{1}{|c|}{336}&0.413&0.426&\underline{0.325}&\underline{0.372}&0.336&0.375&0.176&0.266&0.098&\underline{0.225}&\textbf{0.160}&\textbf{0.211}&\textbf{0.130}&\textbf{0.227}&\textbf{0.365}&\textbf{0.267}&\\
            &\multicolumn{1}{|c|}{720}&0.404&0.432&0.368&0.393&0.332&0.382&0.191&0.277&0.118&0.253&\underline{0.169}&\underline{0.222}&\underline{0.135}&\underline{0.232}&\textbf{0.371}&\textbf{0.275}&\\
            \hline
    \end{tabular}}
    
    \label{second result}
\end{table*}

We not only perform long-term time series prediction tasks but also compare the performance of short-term prediction, and the results are shown in Table \ref{second result}. It can be seen that WaveTS performs equally well in short-term prediction tasks, with good predictive performance on almost all datasets and as the input sequence length increases, the performance is further enhanced. This further indicates that WaveTS can capture effective information in large historical time steps and utilize it, thanks to the decoupling of interference information by DWT. Due to the similar results of WaveTS-B and WaveTS-M in short-term prediction experiments, we replaced them with WaveTS.

\subsubsection{Efficiency comparison}
\label{subsub3}
Evaluate the efficiency of WaveTS based on the number of parameters and Multiply-Accumulate Operations (MACs) \cite{nahmias2019photonic}. Figure \ref{Bubble Diagram} shows the parameter quantities and MACs of different models. The number of parameters indicates the size of the model, while MACs represent the computational requirements for model training. The larger the number of parameters, the more cumbersome the model is, and the larger the MACs, the greater the computational power consumed.
Considering the current large number of Transformer models, which feature hundreds of millions of parameters and long training times, even optimized models have millions of parameters and require lengthy training periods. As shown in Table \ref{Efficiency}, in terms of efficiency, WaveTS performs well, with parameters and MACs comparable to the most efficient linear model. In the subsequent ablation experiments, we verified that WaveTS only loses a small amount of performance in eliminating high-frequency linear layers. By using only low-frequency linear layers, only $50\%$ of the parameter count is used while retaining most of the predictive performance. So we have constructed a single path WaveTS-S that only uses low-frequency information for modeling, further compressing model parameters.

\begin{table*}[ht]
\centering

\caption{The parameter quantity, MACs, and average time required to train an epoch for WaveTS and other mainstream models on the electricity dataset, with a fixed input length of $L=720$ and prediction length of $S=96 $. The single path in parentheses indicates using only low-frequency paths for prediction.}
\resizebox{\textwidth}{!}{
\begin{tabular}{c|ccc|cccccccc}
	\toprule
	Attribute & \textbf{WaveTS-B} &\textbf{WaveTS-S} &\textbf{WaveTS-M} & FITS  & DLinear & iTransformer & Informer  & CrossGNN & Reformer & PatchTST &  FreTS \\
	\midrule
	Parameters  & {69K} &\textbf{40K}&157k & \underline{43K}  & 138K & 304K & 704K  & 730K & 1.1M & 1.3M  & 23M \\
	MACs & {0.71G} &\textbf{0.36G} &1.88G& \underline{0.44G} & 1.42G  & 3.11G & 6.42G  & 9.56G & 8.64G & 9.8G  & 30.32G \\
	Epoch time  & \underline{15.05s} &\textbf{12.11s}&17.65 & 19.93  & {16.43s} & 19.12 & 55.97 & 124.81 & 55.44 & 282.27  & 310.22s\\
	Infer time  & {0.60ms} &\textbf{0.35ms}&0.72ms & 0.79ms  & \underline{0.43ms} & 1.2ms & 2.5ms & 3.4ms & 2.7ms & 1.3ms  & 2.8ms\\
	\bottomrule
	\end{tabular}}
	\label{Efficiency}
\end{table*}

Evaluate the efficiency of WaveTS series models in terms of training time, which is measured as the average training time for each epoch. In most cases, the WaveTS series demonstrates a leading training speed. This efficiency is attributed to using filter banks in DWT to obtain information in different frequency bands, without the need for complex transformation operations.

\subsubsection{Result visualization}

\begin{figure*}[!ht]
    \centering
    \subfloat[Exchange]{%
        \includegraphics[width=0.22\textwidth]{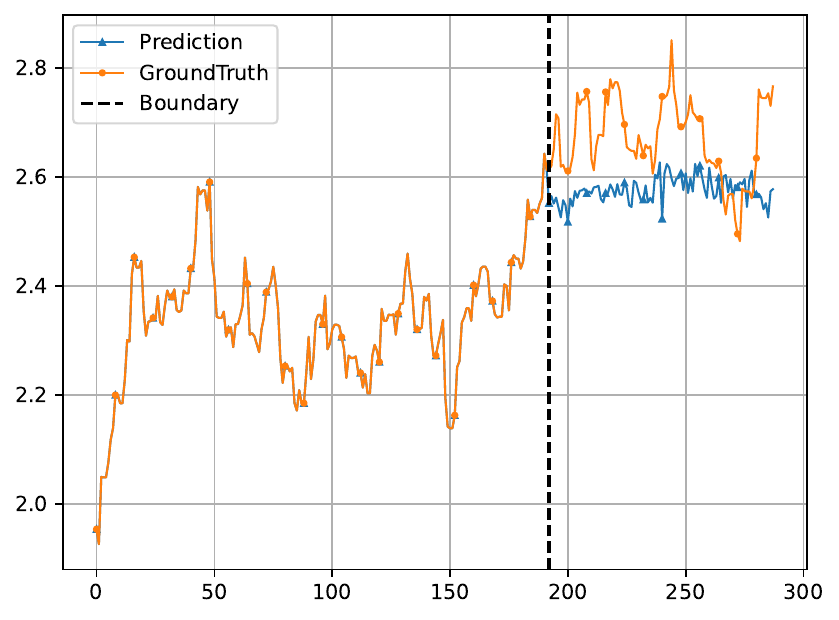}
    }\hfill  
    \subfloat[Electricity]{%
        \includegraphics[width=0.22\textwidth]{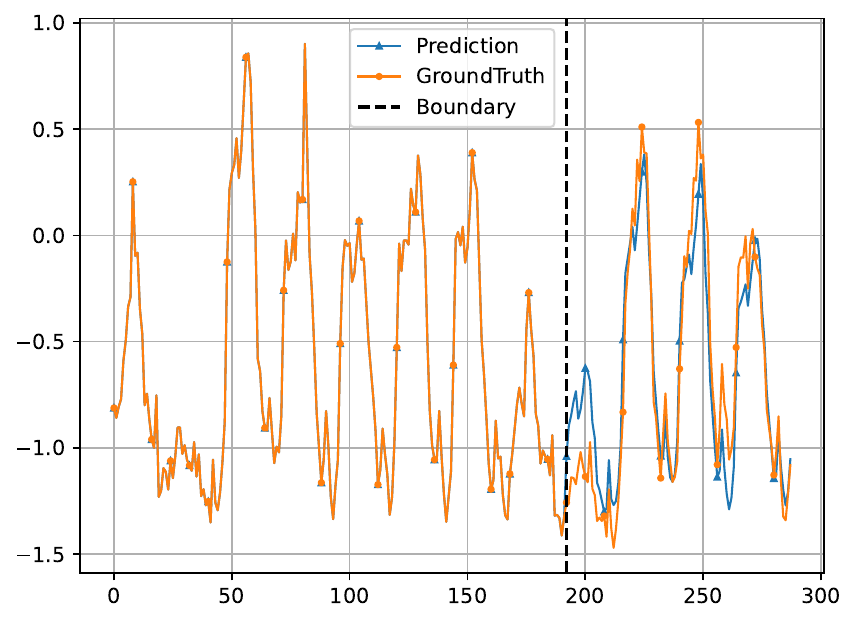}
    }\hfill
    \subfloat[Traffic]{%
        \includegraphics[width=0.22\textwidth]{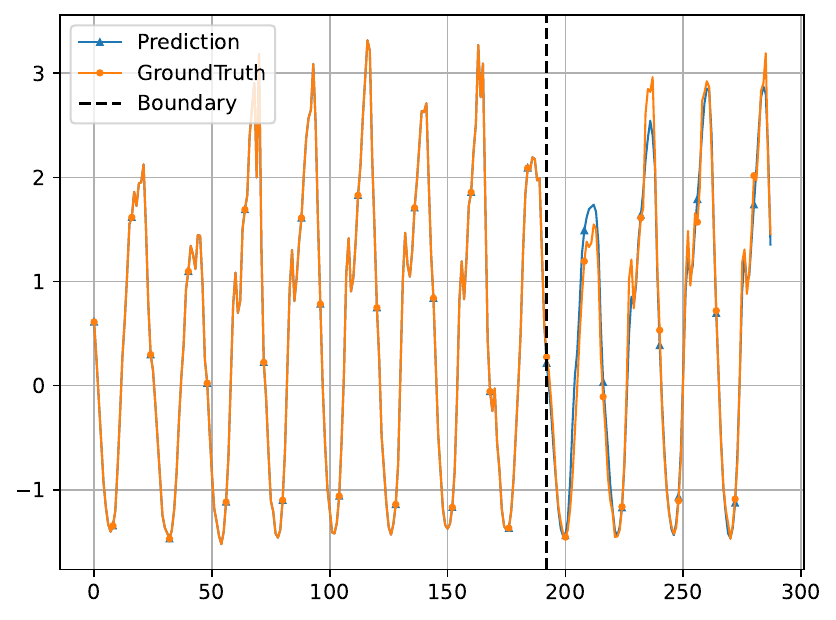}
    }\hfill
    \subfloat[Weather]{%
        \includegraphics[width=0.22\textwidth]{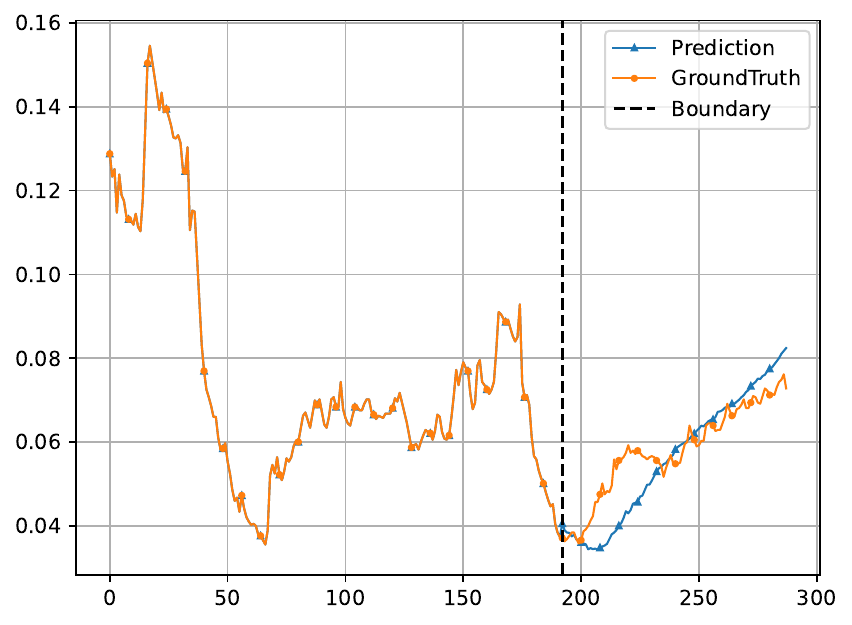}
    }

    \subfloat[ETTh1]{%
        \includegraphics[width=0.22\textwidth]{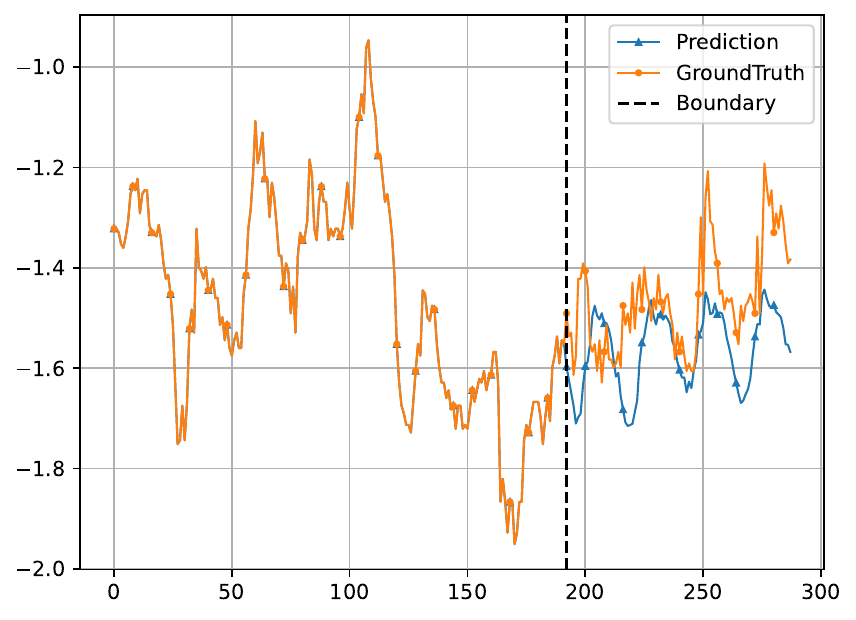}
    }\hfill
    \subfloat[ETTh2]{%
        \includegraphics[width=0.22\textwidth]{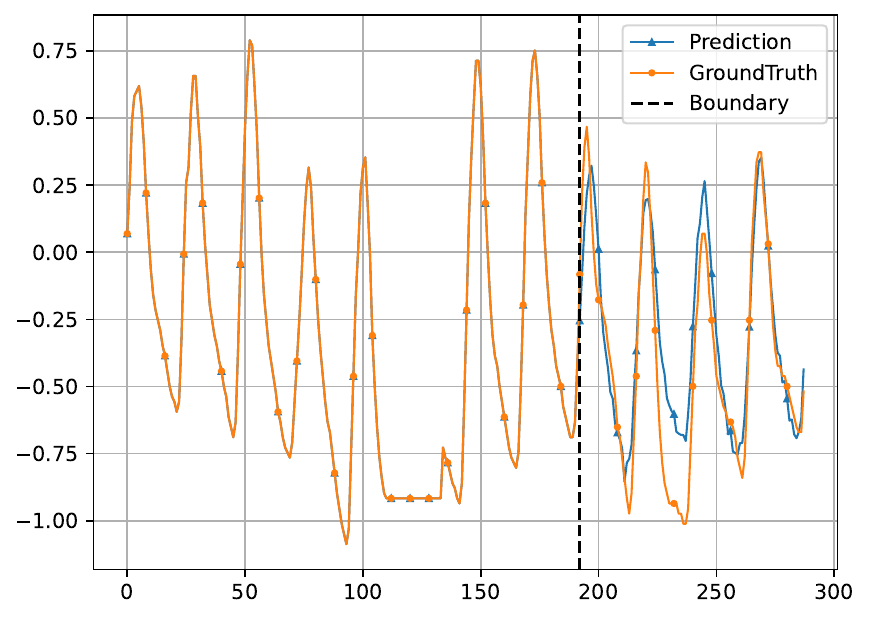}
    }\hfill
    \subfloat[ETTm1]{%
        \includegraphics[width=0.22\textwidth]{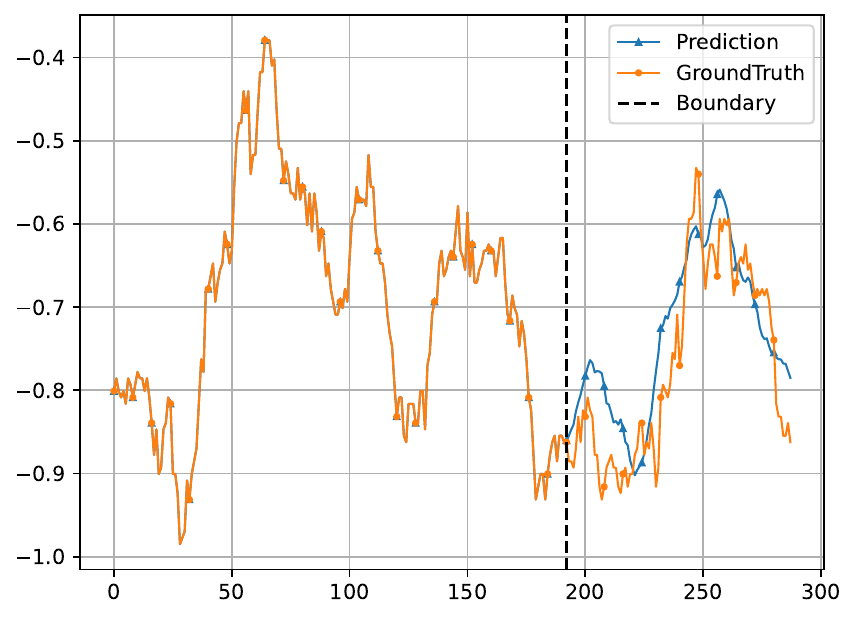}
    }\hfill
    \subfloat[ETTm2]{%
        \includegraphics[width=0.22\textwidth]{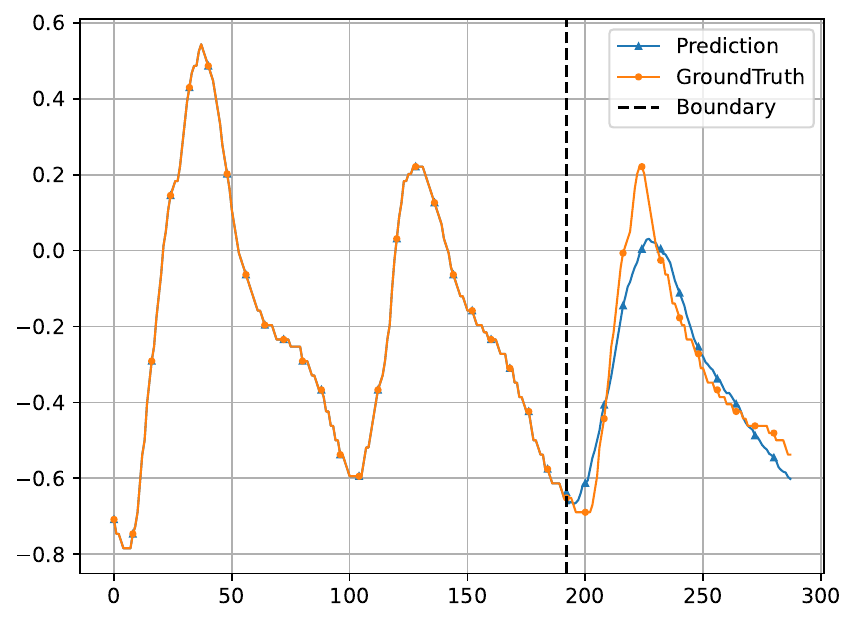}
    }
    \caption{Visualization of results with input length L = 720 and fixed prediction length S = 96.}
    \label{more_results}
\end{figure*}

As shown in Figures \ref{more_results}, we provide a comprehensive visual representation of WaveTS-B’s performance across all datasets. These visualizations offer readers a clear and intuitive way to assess the comparative advantages of our models. The results emphasize the model’s ability to effectively capture non-stationary features, which enhances its capacity to identify complex temporal patterns and improve predictive accuracy. This showcases the robustness of the WaveTS series models in handling datasets with high variability and evolving trends, underscoring their expressiveness and adaptability in a wide range of time series forecasting applications.

\section{Analysis}
Detailed analysis is provided in the Appendix~\ref{analysis}, including ablation experiments, wavelet function analysis, learnable parameter analysis, input length analysis, and visualization analysis.

\section{Conclusion}
\label{Conclusion}
Our foundational model, WaveTS-B, leverages wavelet transformations to enhance predictive accuracy while maintaining computational efficiency. By incorporating a wavelet-driven linear layer, the model simplifies its architecture and eliminates the need for complex inverse transformations. This design significantly reduces the number of model parameters and minimizes computational demands, thus improving efficiency and facilitating the processing of large datasets.
Building on this, we introduce the WaveTS-M model, specifically optimized for multi-channel datasets. This model utilizes a MoE for channel clustering, which enhances performance on multi-channel datasets and achieves higher efficiency compared to larger-scale models. The optimization leverages the inherent correlation characteristics of the data, enabling more effective data processing.
Both models have undergone rigorous theoretical analysis and comprehensive experimental validation. These studies confirm the substantial potential of wavelet transform and MoE techniques in advancing the field of time series prediction, demonstrating improvements in both accuracy and efficiency.

\noindent\textbf{Strengths and Limitations:} The WaveTS series models excel in real-time applications, such as smart grid energy forecasting, due to their efficient and streamlined design, which ensures quick response times. Despite its simplicity, WaveTS consistently demonstrates superior predictive accuracy, surpassing traditional deep learning approaches by approximately 12\% on mean squared error metrics in datasets like ETTh2. However, the model encounters challenges when dealing with data characterized by high levels of random fluctuations or non-periodic features, such as those found in financial markets. In these contexts, WaveTS may occasionally miss short-term spikes or drops, highlighting areas where further refinement is needed to enhance its adaptability and predictive reliability.

\noindent\textbf{Future Work:} To enhance the model's generalization and robustness, future research will explore a broader array of wavelet bases that may offer better symmetry and reconstruction properties for specific applications. This will be particularly targeted at improving the model's performance in handling high-frequency fluctuations observed in financial and meteorological data. Additionally, given the potential applications of WaveTS series models in the energy and transportation sectors, we aim to extend its use to other practical scenarios such as climate change forecasting and smart manufacturing process control. These efforts will help validate the model's practicality and effectiveness in real-world settings that demand high-frequency data analysis.

\bibliography{reference}
\bibliographystyle{unsrt}

\appendix
\newpage
\section{Algorithm}
\label{algorithm}
\begin{algorithm}
	\caption{WaveTS - Overall Architecture}
	
	\begin{algorithmic}[1]
		\Require historical observation data $X=\left \{ x_1,\dots ,x_l  \right \} \in \mathbb{R} ^{L\times N}$; input length $L$; predicted length $S$; variates number $N$; LF~(Wavelet low-pass filter), HF~(Wavelet high-pass filter).
		\State $\rhd $ Reversible instance normalization.
		\State $X \gets RevIN(X)$ \Comment {$X \in \mathbb{R}^{L \times N}$}
		\State $\rhd $ Low-pass filter coefficients $=\begin{bmatrix} \alpha &\alpha  \end{bmatrix}$.
		\State $X_A \gets LF(X)$  \Comment {$X_A \in \mathbb{R}^{\frac{L}{2} \times N}$}
		\State $\rhd $ High-pass filter coefficients $=\begin{bmatrix} \alpha &-\alpha  \end{bmatrix}$.
		\State $X_D \gets HF(X)$  \Comment  $X_D \in \mathbb{R}^{\frac{L}{2} \times N}$	
		\State $X_A,X_D \gets X_A^\top, X_D^\top$ \Comment $X_A \in \mathbb{R}^{N \times \frac{L}{2}}$, $X_D \in \mathbb{R}^{N \times \frac{L}{2}}$
        \State \textbf{WaveTS-M}:
		\State $\rhd $ Applying MoE and linear projection $\frac{L}{2} \to S$.
		\State $Y_A, Y_D = MoE(X_A), Linear(X_D) $ \Comment $Y_A,Y_D \in \mathbb{R}^{N \times S}$
        \State \textbf{WaveTS-B}:
        \State $\rhd $ Applying linear projection $\frac{L}{2} \to 2S$.
		\State $Y_A, Y_D = MLP(X_A), Linear(X_D) $ \Comment $Y_A,Y_D \in \mathbb{R}^{N \times S}$
        \State $Y \gets Y_A + \delta Y_D$ \Comment $Y \in \mathbb{R}^{N \times S}$
		\State $\rhd$ Apply the inverse of reversible instance normalization.
		\State $Y \gets iRevIN(Y^\top)$ \Comment $Y \in \mathbb{R}^{S \times N}$
		\State \Return $Y$ \Comment Return the prediction result $Y$
	\end{algorithmic}
\end{algorithm}

\section{Datasets}
\label{database}
\begin{table}[ht]
    \caption{Summary of eight benchmarks.}
    \scriptsize
    \renewcommand{\arraystretch}{1.5}
	\centering
	\resizebox{\linewidth}{!}{\setlength{\tabcolsep}{0.35mm}{
		\begin{tabular}{c|cccccc} \hline
			Datasets       & ETTh1$\&$ETTh2       &ETTm1$\&$ETTm2        & Traffic     & Electricity  & Exchange-Rate & Weather            \\ 
			\hline
			variable number       & 7          & 7            & 862         & 321          & 8             & 21                \\
			Length         & 17,420     & 69,680       & 17,544      & 26,304       & 7,588         & 52,696            \\
			Step    & 1hour      & 5min         & 1hour       & 1hour        & 1day          & 10min           \\ 
			\hline
	\end{tabular}}}
	
\end{table}

The specific characteristics of each dataset are described in detail below:
\begin{enumerate}[label=\arabic*)]
	\item \textbf{ETT (ETTh1, ETTh2, ETTm1, ETTm2)} consists of two hourly-level datasets (ETTh) and two 15-minute-level datasets (ETTm). Each of them contains seven oil and load features of electricity transformers from July 2016 to July 2018.
	\item \textbf{Traffic} describes hourly road occupancy rates measured by 862 sensors on San Francisco Bay area freeways from 2015 to 2016.
	\item \textbf{Electricity} contains the hourly electricity consumption of 321 clients from 2012 to 2014.
	\item \textbf{Exchange-rate} collects the daily exchange rates of 8 countries from 1990 to 2016.
	\item \textbf{Weather} includes 21 indicators of weather, such as air temperature, and humidity. Its data is recorded every 10 min for 2020 in Germany.
\end{enumerate}



\section{Model analysis}
\label{analysis}

\subsection{Ablation experiment of decomposition}
\label{Decomposition}
\begin{table*}[th]
	\caption{Ablation experiment of low-frequency (LF) and high-frequency (HF) linear layers in the first three lines. And ablation experiment of using wavelet inverse transform and direct use of domain transformer in lines one and four. (Input length $L=720$ and prediction length $S=96$)}
	\centering
    \scriptsize

		\begin{tabular}{c|c|c|c|c|c|c|c|c|c}
			\hline
			\multicolumn{1}{c|}{Models}&\multicolumn{1}{c|}{Metric}&Exchange&Weather&Electricity&Traffic&ETTh1&ETTh2&ETTm1&ETTm2\\
			\hline
			\multirow{2}{*}{WaveTS-B}&\multicolumn{1}{c|}{MSE}&\textbf{0.083}&\textbf{0.167}&\textbf{0.133}&\textbf{0.377}&\textbf{0.377}&\textbf{0.270}&\textbf{0.300}&\textbf{0.161}\\
			&\multicolumn{1}{c|}{MAE}&\textbf{0.203}&\textbf{0.220}&\textbf{0.228}&\textbf{0.265}&\textbf{0.400}&\textbf{0.335}&\textbf{0.347}&\textbf{0.251}\\
			\hline
			\multirow{2}{*}{WaveTS-LF}&\multicolumn{1}{c|}{MSE}&0.090&0.168&0.145&0.398&0.430&0.271&0.351&0.182\\
			&\multicolumn{1}{c|}{MAE}&0.212&0.221&0.248&0.282&0.452&0.336&0.372&0.265\\
			\hline
			\multirow{2}{*}{WaveTS-HF}&\multicolumn{1}{c|}{MAE}&0.532&0.284&0.279&0.551&0.548&0.379&0.406&0.296\\
			&\multicolumn{1}{c|}{MSE}&0.572&0.328&0.378&0.418&0.549&0.424&0.427&0.355\\
			\hline
			\multirow{2}{*}{WaveTS-I}&\multicolumn{1}{c|}{MAE}&0.092&0.175&0.135&0.388&0.394&0.298&0.312&0.175 \\
			&\multicolumn{1}{c|}{MSE}&0.211&0.248&0.234&0.319&0.526&0.354&0.377&0.325\\
			\hline
	\end{tabular}
	\label{Ablation frequency}
\end{table*}

Wavelet decomposition separates time series into high-frequency and low-frequency components, where the low-frequency part captures the primary sequence information, and the high-frequency part captures local details. However, the inverse transformation introduces some reconstruction errors. To validate this, we removed the linear layer from the high-frequency and low-frequency parts of the model, respectively, and replaced the domain transformer with the wavelet inverse transform to create a new model, WaveTS-I. As shown in Table \ref{Ablation frequency}, the model using only the high-frequency part fails to fit the data accurately, resulting in a significant increase in both MSE and MAE, reaching several times the values of the original model. In contrast, the model using only the low-frequency part and WaveTS-I performs slightly worse than the original model, which aligns with our expectations.

\subsection{Analysis of different wavelets}
DWT is a crucial tool in signal analysis that effectively captures both time and frequency information of signals through multi-scale decomposition. Here we introduce four common types of DWT along with their basic filtering coefficients:

\begin{itemize}
	\item Haar wavelet transform~\cite{singh2021advancement} it is computationally straightforward, allowing for quick wavelet decomposition and reconstruction. Its filter coefficient is $\begin{bmatrix} \frac{1}{\sqrt{2} } &\frac{1}{\sqrt{2} } \\ \frac{1}{\sqrt{2} } &-\frac{1}{\sqrt{2} }   \end{bmatrix}$.
	\item Daubechies wavelet transform~\cite{manimaran2009multiresolution}, Invented by Ingrid Daubechies, this wavelet is compactly supported and highly smooth. Daubechies wavelets provide better smoothness and longer support. Its filter coefficient is (D4 example) $\begin{bmatrix} \frac{\sqrt{2} + 2}{4\sqrt{2}}, \frac{3\sqrt{2} + 2}{4\sqrt{2}}, \frac{2 - \sqrt{2}}{4\sqrt{2}}, -\frac{2 - 3\sqrt{2}}{4\sqrt{2}} \\ -\frac{2 - 3\sqrt{2}}{4\sqrt{2}}, -\frac{2 - \sqrt{2}}{4\sqrt{2}}, \frac{3\sqrt{2} + 2}{4\sqrt{2}}, -\frac{\sqrt{2} + 2}{4\sqrt{2}}   \end{bmatrix}$.
    \item Coiflets wavelet transform~\cite{zaeni2018application} offer high symmetry and longer filter lengths, making them particularly suitable for processing signals with smooth characteristics. Its filter coefficient is (Coif1 example) 
	
	$\begin{bmatrix} -\frac{49}{3270\sqrt{2}}, -\frac{91}{1250\sqrt{2}}, \frac{961}{2500\sqrt{2}}, \frac{4261}{5000\sqrt{2}}, \frac{1689}{5000\sqrt{2}}, -\frac{91}{1250\sqrt{2}} \\ \frac{91}{1250\sqrt{2}}, \frac{1689}{5000\sqrt{2}}, -\frac{4261}{5000\sqrt{2}}, \frac{961}{2500\sqrt{2}}, \frac{91}{1250\sqrt{2}}, -\frac{49}{3270\sqrt{2}}   \end{bmatrix}$.
\end{itemize}

\begin{table*}[th]
	\caption{Different wavelet selection on Electricity dataset. The best result is highlighted in \textbf{bold}. (Input length $T=720$ and prediction length $S\in\left\{96, 192, 336, 720\right\}$)}
	\centering
	\resizebox{\textwidth}{!}{\setlength{\tabcolsep}{3mm}{
			\begin{tabular}{c|ccc|ccc|ccc|ccc}
				\hline
				\multicolumn{1}{c|}{Wavelet Selection}  & \multicolumn{3}{c|}{96} & \multicolumn{3}{c|}{192}  & \multicolumn{3}{c|}{336} & \multicolumn{3}{c}{720} \\ 
				\hline
				\multicolumn{1}{c|}{Metric} & \multicolumn{1}{c}{MSE} & \multicolumn{1}{c}{MAE} & \multicolumn{1}{c|}{Time} & \multicolumn{1}{c}{MSE} & \multicolumn{1}{c}{MAE} & \multicolumn{1}{c|}{Time}& \multicolumn{1}{c}{MSE} & \multicolumn{1}{c}{MAE} & \multicolumn{1}{c|}{Time}& \multicolumn{1}{c}{MSE} & \multicolumn{1}{c}{MAE} & \multicolumn{1}{c}{Time} \\ 
				\hline
				
				\multirow{1}{*}{D4}&0.134 & 0.230 & 17.92s & \textbf{0.148} & \textbf{0.242} & 19.15s & \textbf{0.164} & \textbf{0.258} & 21.49s & 0.203 & 0.291 & 27.83s  \\
				
				\multirow{1}{*}{Sym4}&\textbf{0.133} & 0.229 & 19.04s & 0.149 & 0.244 & 21.30s & \textbf{0.164} & 0.259 & 23.75s & \textbf{0.202} & \textbf{0.290} & 30.68s  \\
				\multirow{1}{*}{Coif1}&0.134 & 0.230 & 16.47s & \textbf{0.148} & 0.243 & 18.52s & \textbf{0.164} & 0.259 & 21.05s & 0.204 & 0.292 & 27.27s  \\
				
				\multirow{1}{*}{Fourier}&0.144 & 0.330 & 19.19s & 0.152 & 0.246 & 24.41s & 0.169 & 0.263 & 28.97s & 0.210 & 0.301 & 33.82s  \\
				\hline
				\multirow{1}{*}{\textbf{WaveTS-B}}&\textbf{0.133} & \textbf{0.228} & \textbf{15.67s} & \textbf{0.148} & \textbf{0.242} & \textbf{16.88s} & \textbf{0.164} & \textbf{0.258} & \textbf{19.63s} & 0.203 & 0.291 & \textbf{26.59s}  \\
				\hline
	\end{tabular}}}
	\label{Different wavelet}
\end{table*}

In addition to evaluating the impact of decomposition levels, we conducted a series of experiments on the electrical energy dataset to assess how different wavelet functions influence both model performance and computational efficiency. The results are detailed in Table \ref{Different wavelet}, which compares the performance of WaveTS-B under various wavelet functions.
It is evident from the results that WaveTS-B achieves superior predictive accuracy when using more complex wavelet functions, such as the Daubechies or Coiflet families, which are known for their ability to capture intricate patterns in the data across multiple frequency bands. However, the Haar wavelet function, due to its simple structure and binary filter coefficients, offers clear advantages in terms of computational efficiency. This is particularly relevant in scenarios where speed and resource constraints are critical, as Haar’s reduced computational complexity minimizes the number of required operations.
The trade-off between performance and efficiency is evident: while more sophisticated wavelet functions can improve predictive accuracy by better capturing the nuances in the time series, they come at the cost of increased computational overhead.

\subsection{Analysis of learnable parameters}
\begin{table*}[th]
	\caption{The best result is highlighted in \textbf{bold}. Input length $T=720$ and prediction length $S\in\left\{96, 192, 336, 720\right\}$.}
	\renewcommand{\arraystretch}{1.4}
	\centering
	\resizebox{\textwidth}{!}{
			\begin{tabular}{c|c|cccc|cccc|cccc|cccc}
				\hline
				\multicolumn{2}{c|}{Dataset}  & \multicolumn{4}{c|}{ETTh1} & \multicolumn{4}{c|}{ETTh2}  & \multicolumn{4}{c|}{ETTm1} & \multicolumn{4}{c}{ETTm2} \\ 
				\hline
				\multicolumn{2}{c|}{Horizon} & \multicolumn{1}{c}{96} & \multicolumn{1}{c}{192} & \multicolumn{1}{c}{336} & \multicolumn{1}{c|}{720} & \multicolumn{1}{c}{96} & \multicolumn{1}{c}{192} & \multicolumn{1}{c}{336} & \multicolumn{1}{c|}{720} & \multicolumn{1}{c}{96} & \multicolumn{1}{c}{192} & \multicolumn{1}{c}{336} & \multicolumn{1}{c|}{720} & \multicolumn{1}{c}{96} & \multicolumn{1}{c}{192} & \multicolumn{1}{c}{336} & \multicolumn{1}{c}{720} \\ 
				\hline
				\multirow{2}{*}{Weight \ding{51}}&\multicolumn{1}{c|}{MSE} & \textbf{0.377} & \textbf{0.412} & \textbf{0.438} & \textbf{0.455} & 0.276 & \textbf{0.333} & \textbf{0.357} & \textbf{0.383} & \textbf{0.309} & \textbf{0.342} & \textbf{0.367} & \textbf{0.418} & 0.163 & \textbf{0.215} & \textbf{0.268} & \textbf{0.350} \\
				&\multicolumn{1}{c|}{MAE}& \textbf{0.401} & \textbf{0.424} & \textbf{0.443} & \textbf{0.475} & 0.337 & \textbf{0.375} & \textbf{0.399} & \textbf{0.428} & \textbf{0.351} & \textbf{0.371} & \textbf{0.385} & \textbf{0.413} & \textbf{0.253} & \textbf{0.288} & \textbf{0.324} & \textbf{0.377}\\
				\hline
				\multirow{2}{*}{Weight \ding{55}}&\multicolumn{1}{c|}{MSE} & 0.38 & 0.421 & 0.452 & 0.470 & \textbf{0.270} & 0.336 & 0.358 & 0.385 & 0.309 & 0.343 & 0.371 & 0.421 & \textbf{0.162} & 0.216 & 0.270 & 0.350 \\
				&\multicolumn{1}{c|}{MAE}& 0.403 & 0.431 & 0.450 & 0.480 & \textbf{0.335} & 0.378 & 0.399 & 0.430 &0.352& 0.372 & 0.388 & 0.416 &  0.253 & 0.290 & 0.327 & 0.378\\
				\hline
	\end{tabular}}
	
	\label{Ablation weight}
\end{table*}
Wavelet decomposition divides time series into different frequency bands, with time-series data typically comprising several main frequencies. Following wavelet decomposition, we introduce learnable parameters for high-frequency bands to adaptively adjust the weights of these bands. Consequently, we conducted ablation experiments using learnable parameters to investigate the impact of different frequency band ratios. As shown in Table \ref{Ablation weight}, the results indicate that learnable parameters can autonomously allocate weights to the frequency bands, thereby reducing errors and enhancing prediction accuracy.

\subsection{Analysis of stability}
We report the standard deviation of WaveTS performance under five runs with different random seeds in Table \ref{Stability reuslt}, which exhibits that the performance of WaveTS is stable.
\label{stability}
\begin{table}[th]
	\caption{Robustness of WaveTS performance. The results are obtained from five random seeds. (Input length $L=720$ and prediction length $T\in\left\{96, 192, 336, 720\right\}$)}
	\scriptsize
	\renewcommand{\arraystretch}{1.4}
	\centering
	\begin{subtable}[t]{\textwidth}
		\centering
		\resizebox{\textwidth}{!}{\setlength{\tabcolsep}{0.35mm}{
				\begin{tabular}{c|cccc|cccc|cccc|cccc}
					\toprule
					Dataset  & \multicolumn{4}{c|}{Exchange}& \multicolumn{4}{c|}{Weather} & \multicolumn{4}{c|}{Electricity}  & \multicolumn{4}{c}{Traffic}  \\ \hline
					Horizon & \multicolumn{2}{c}{MSE} & \multicolumn{2}{c|}{MAE} & \multicolumn{2}{c}{MSE} & \multicolumn{2}{c|}{MAE}& \multicolumn{2}{c}{MSE} & \multicolumn{2}{c|}{MAE}& \multicolumn{2}{c}{MSE} & \multicolumn{2}{c}{MAE} \\ \midrule
					96 & \multicolumn{2}{c}{0.083$\pm$0.002} & \multicolumn{2}{c|}{0.203$\pm$0.001}& \multicolumn{2}{c}{0.167$\pm$0.001} & \multicolumn{2}{c|}{0.220$\pm$0.000}& \multicolumn{2}{c}{0.133$\pm$0.001} & \multicolumn{2}{c|}{0.228$\pm$0.001}& \multicolumn{2}{c}{0.377$\pm$0.002} & \multicolumn{2}{c|}{0.265$\pm$0.001} \\
					192 & \multicolumn{2}{c}{0.174$\pm$0.001} & \multicolumn{2}{c|}{0.297$\pm$0.001}& \multicolumn{2}{c}{0.210$\pm$0.001} & \multicolumn{2}{c|}{0.257$\pm$0.002}& \multicolumn{2}{c}{0.148$\pm$0.001} & \multicolumn{2}{c|}{0.242$\pm$0.001}& \multicolumn{2}{c}{0.390$\pm$0.001} & \multicolumn{2}{c|}{0.272$\pm$0.002} \\
					336 & \multicolumn{2}{c}{0.338$\pm$0.001} & \multicolumn{2}{c|}{0.421$\pm$0.002}& \multicolumn{2}{c}{0.256$\pm$0.002} & \multicolumn{2}{c|}{0.293$\pm$0.001}& \multicolumn{2}{c}{0.164$\pm$0.001} & \multicolumn{2}{c|}{0.258$\pm$0.000}& \multicolumn{2}{c}{0.403$\pm$0.002} & \multicolumn{2}{c|}{0.275$\pm$0.001}\\
					720 & \multicolumn{2}{c}{1.025$\pm$0.002} & \multicolumn{2}{c|}{0.762$\pm$0.001}& \multicolumn{2}{c}{0.319$\pm$0.002} & \multicolumn{2}{c|}{0.338$\pm$0.001}& \multicolumn{2}{c}{0.203$\pm$0.001} & \multicolumn{2}{c|}{0.291$\pm$0.002}& \multicolumn{2}{c}{0.442$\pm$0.001} & \multicolumn{2}{c|}{0.294$\pm$0.001} \\
					 \bottomrule
		\end{tabular}}}
	\end{subtable}
	\hfill
	\begin{subtable}[t]{\textwidth}
		\centering
		\resizebox{\textwidth}{!}{\setlength{\tabcolsep}{0.35mm}{
				\begin{tabular}{c|cccc|cccc|cccc|cccc}
					\toprule
					Dataset  & \multicolumn{4}{c|}{ETT-h1} & \multicolumn{4}{c|}{ETT-h2}  & \multicolumn{4}{c|}{ETT-m1} & \multicolumn{4}{c}{ETT-m2} \\ \hline
					Horizon & \multicolumn{2}{c}{MSE} & \multicolumn{2}{c|}{MAE} & \multicolumn{2}{c}{MSE} & \multicolumn{2}{c|}{MAE}& \multicolumn{2}{c}{MSE} & \multicolumn{2}{c|}{MAE}& \multicolumn{2}{c}{MSE} & \multicolumn{2}{c}{MAE} \\ \midrule
					96 & \multicolumn{2}{c}{0.377$\pm$0.001} & \multicolumn{2}{c|}{0.400$\pm$0.001} & \multicolumn{2}{c}{0.270$\pm$0.002} & \multicolumn{2}{c|}{0.335$\pm$0.001}& \multicolumn{2}{c}{0.300$\pm$0.002} & \multicolumn{2}{c|}{0.347$\pm$0.001}& \multicolumn{2}{c}{0.161$\pm$0.000} & \multicolumn{2}{c}{0.251$\pm$0.001} \\
					192 & \multicolumn{2}{c}{0.421$\pm$0.001} & \multicolumn{2}{c|}{0.427$\pm$0.001} & \multicolumn{2}{c}{0.336$\pm$0.002} & \multicolumn{2}{c|}{0.378$\pm$0.002}& \multicolumn{2}{c}{0.337$\pm$0.002} & \multicolumn{2}{c|}{0.368$\pm$0.001}& \multicolumn{2}{c}{0.216$\pm$0.001} & \multicolumn{2}{c}{0.290$\pm$0.001} \\
					336 & \multicolumn{2}{c}{0.452$\pm$0.001} & \multicolumn{2}{c|}{0.446$\pm$0.001} & \multicolumn{2}{c}{0.358$\pm$0.002} & \multicolumn{2}{c|}{0.399$\pm$0.003}& \multicolumn{2}{c}{0.371$\pm$0.001} & \multicolumn{2}{c|}{0.388$\pm$0.001}& \multicolumn{2}{c}{0.270$\pm$0.004} & \multicolumn{2}{c}{0.327$\pm$0.001}\\
					720 & \multicolumn{2}{c}{0.470$\pm$0.001} & \multicolumn{2}{c|}{0.480$\pm$0.001} & \multicolumn{2}{c}{0.385$\pm$0.001} & \multicolumn{2}{c|}{0.430$\pm$0.001}& \multicolumn{2}{c}{0.417$\pm$0.005} & \multicolumn{2}{c|}{0.416$\pm$0.001}& \multicolumn{2}{c}{0.350$\pm$0.001} & \multicolumn{2}{c}{0.378$\pm$0.006} \\
					 \bottomrule
		\end{tabular}}}
	\end{subtable}
	
	\label{Stability reuslt}
\end{table}

\subsection{Increasing input length}
\begin{figure*}[th]
	\centering
	\includegraphics[width=\textwidth]{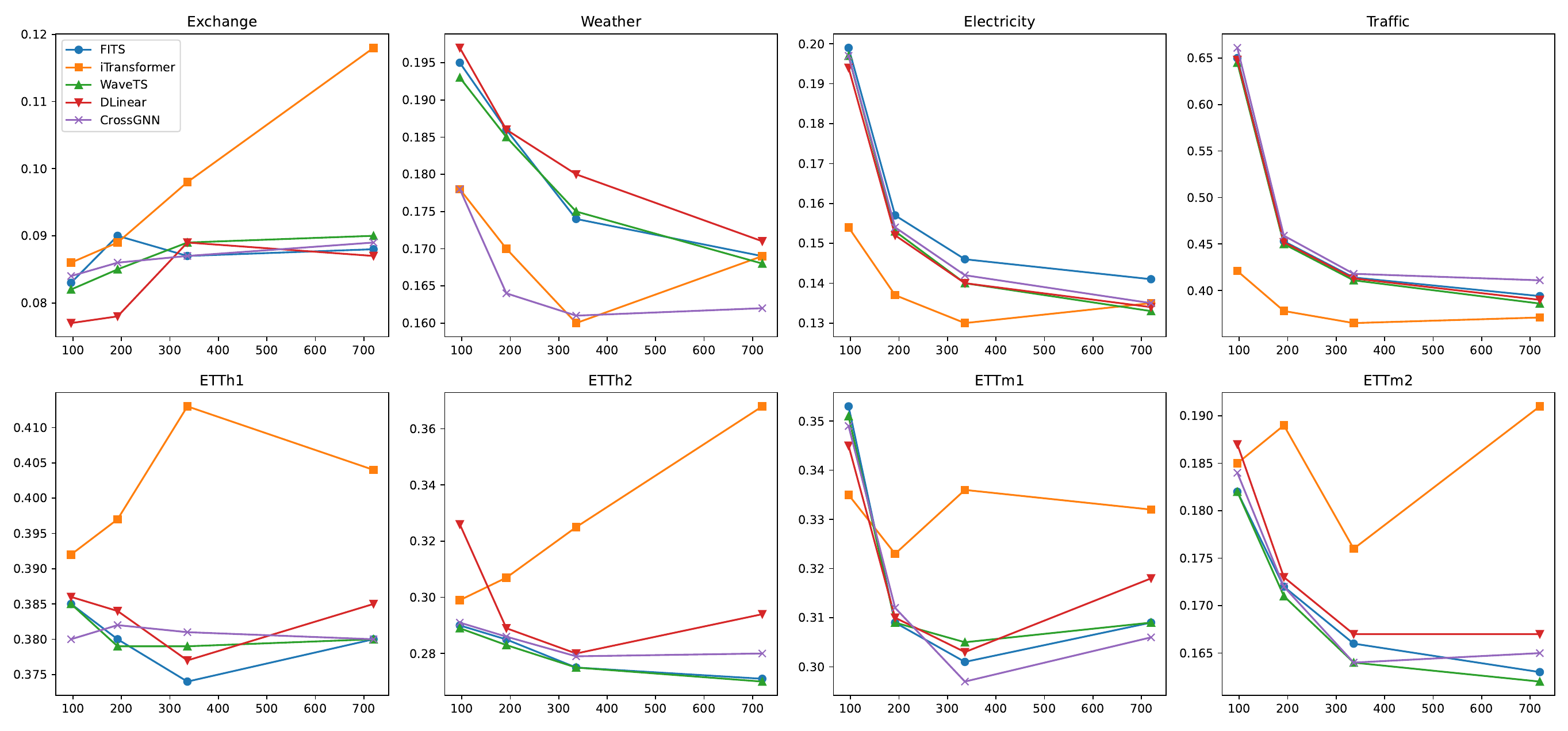}
	\caption{The performance of increasing input length $L\in\left\{96, 192, 336, 720\right\} $ and fixed prediction length $S = 96$. }
	\label{ablation input length}
\end{figure*}

As illustrated in Figure \ref{ablation input length}, the predictive performance of WaveTS series models improves progressively with the increase in historical time steps. This enhancement can be attributed to WaveTS's ability to capture richer temporal information and uncover the underlying distributional patterns inherent in the time series. By leveraging longer historical sequences, the model can effectively disentangle complex trends and seasonality, leading to more accurate forecasts.
However, in datasets where periodicity is weak or non-existent, and the data is heavily influenced by random noise—such as the Exchange dataset—a different trend is observed. In these cases, most models, including WaveTS series models, exhibit a rising trend in prediction error as the length of the historical input increases. This phenomenon is primarily driven by the cumulative propagation of errors, which is particularly pronounced in autoregressive models. As the prediction horizon extends, small errors in earlier forecasts compound over time, resulting in a noticeable degradation in performance.
This observation highlights the importance of balancing historical input length with the characteristics of the data. While longer input sequences generally provide more information, they can also exacerbate error propagation in models sensitive to such accumulative effects, especially in non-periodic and noisy datasets.



\end{document}